\documentclass[a4paper]{cas-sc}
\usepackage[numbers]{natbib}
\usepackage{multirow}%
\usepackage{amsmath,amssymb,amsfonts}%
\usepackage{amsthm}%
\usepackage{mathrsfs}%
\usepackage[title]{appendix}%
\usepackage{xcolor}%
\usepackage{textcomp}%
\usepackage{manyfoot}%
\usepackage{booktabs}%
\usepackage{algorithm}%
\usepackage{algorithmicx}%
\usepackage{algpseudocode}%
\usepackage{listings}%
\usepackage{appendix} 
\usepackage{amsmath}
\usepackage{multirow}
\usepackage{balance}
\usepackage{amssymb}
\usepackage{amsmath}
\usepackage{longtable}
\usepackage{subfloat} 
\usepackage[switch]{lineno}
\usepackage{xcolor}
\usepackage{graphicx}
\usepackage{subfig}
\usepackage{bbding}
\usepackage{color}
\usepackage{nicematrix}

\def\tsc#1{\csdef{#1}{\textsc{\lowercase{#1}}\xspace}}
\tsc{WGM}
\tsc{QE}
\tsc{EP}
\tsc{PMS}
\tsc{BEC}
\tsc{DE}

\begin{document}
\let\WriteBookmarks\relax
\def\floatpagepagefraction{1}
\def\textpagefraction{.001}
\shorttitle{MapFusion: A Novel BEV Feature Fusion Network for Multi-modal Map Construction}
\shortauthors{Hao et~al.}

\title [mode = title]{MapFusion: A Novel BEV Feature Fusion Network for Multi-modal Map Construction}

\author[inst1]{Xiaoshuai Hao}
\orcidauthor{0009-0007-4209-6695}{Xiaoshuai Hao}
\ead{xshao@baai.ac.cn}

\author[inst2]{Yunfeng Diao}
\orcidauthor{0000-0002-9455-1510}{Yunfeng Diao}
\cormark[1]
\ead{diaoyunfeng@hfut.edu.cn}

\author[inst11]{Mengchuan Wei}
\ead{mc.wei@samsung.com}

\author[inst11]{Yifan Yang}
\ead{yifan.yang@samsung.com}

\author[inst11]{Peng Hao}
\orcidauthor{0000-0001-6448-3062}{Peng Hao}
\ead{peng1.hao@samsung.com}

\author[inst3]{Rong Yin}
\orcidauthor{0000-0003-1894-7561}{Rong Yin}
\cormark[1]
\ead{yinrong@iie.ac.cn}

\author[inst11]{Hui Zhang}
\orcidauthor{0000-0002-9912-1157}{Hui Zhang}
\ead{hui123.zhang@samsung.com}

\author[inst11]{Weiming Li}
\orcidauthor{0000-0003-4054-5956}{Weiming Li}
\ead{weiming.li@samsung.com}

\author[inst4]{Shu Zhao}
\orcidauthor{0009-0001-2310-9442}{Shu Zhao}
\ead{smz5505@psu.edu}

\author[inst5]{Yu Liu}
\orcidauthor{0000-0003-2211-3535}{Yu Liu}
\ead{yuliu@hfut.edu.cn}

\cortext[inst3,inst4]{Corresponding author.}

\affiliation[inst1]{organization={Beijing Academy of Artificial Intelligence},
            addressline={150 Chengfu Road, Haidian District}, 
            postcode={100083}, 
            state={Beijing},
            country={China}}

 \affiliation[inst2]{organization={ School of Computer Science and Information Engineering, Hefei University of Technology},
             addressline={Shushan District}, 
             postcode={230009}, 
            state={Hefei},
            country={China }}

\affiliation[inst11]{organization={Samsung R\&D Institute China–Beijing},
            addressline={No. 12, Taiyangong Middle Road}, 
            postcode={100028}, 
            state={Beijing},
            country={China}}

\affiliation[inst3]{organization={Institute of Information Engineering, Chinese Academy of Sciences},
            addressline={No. 19, Rd. Shucun}, 
            postcode={100195}, 
            state={Beijing},
            country={China}}

\affiliation[inst4]{organization={Pennsylvania State University},
            addressline={State College}, 
            postcode={16801}, 
            state={Pennsylvania},
            country={United States}}

 \affiliation[inst5]{organization={Department of Biomedical Engineering, Hefei University of Technology},
             addressline={Shushan District}, 
             postcode={230009}, 
            state={Hefei},
            country={China}}

\begin{abstract}
Map construction task plays a vital role in providing precise and comprehensive static environmental information essential for autonomous driving systems.
Primary sensors include cameras and LiDAR, with configurations varying between camera-only, LiDAR-only, or camera-LiDAR fusion, based on cost-performance considerations. 
While fusion-based methods typically perform best, existing approaches often neglect modality interaction and rely on simple fusion strategies, which suffer from the problems of misalignment and information loss.
To address these issues, we propose \textit{MapFusion}, a novel multi-modal Bird's-Eye View (BEV) feature fusion method for map construction.
Specifically, to solve the semantic misalignment problem between camera and LiDAR BEV features, we introduce the Cross-modal Interaction Transform (CIT) module, enabling interaction between two BEV feature spaces and enhancing feature representation through a self-attention mechanism.
Additionally, we propose an effective Dual Dynamic Fusion (DDF) module to adaptively select valuable information from different modalities, which can take full advantage of the inherent information between different modalities. 
Moreover, \textit{MapFusion} is designed to be simple and plug-and-play, easily integrated into existing pipelines. 
We evaluate \textit{MapFusion} on two map construction tasks, including High-definition (HD) map and BEV map segmentation, to show its versatility and effectiveness.
Compared with the state-of-the-art methods, MapFusion achieves 3.6\% and 6.2\% absolute improvements on the
HD map construction and BEV map segmentation tasks on the nuScenes dataset, respectively, demonstrating the
superiority of our approach.

\end{abstract}

\begin{keywords}
BEV Feature Fusion \sep 
Cross-modal Interaction \sep
Dual Dynamic Fusion \sep
Multi-modal Map Construction 
\end{keywords}

\maketitle

\section{Introduction}

Map construction task provides abundant and precise static environmental information of the driving scene, which is vital yet challenging for planning in autonomous driving systems. 
Recently, researchers have focused on two crucial tasks: High-definition (HD) map construction and semantic map construction.
Both tasks increasingly utilize the Bird's Eye View (BEV) representation as an ideal feature space for multi-view perception, thanks to its effectiveness in addressing scale ambiguity and occlusion challenges.
Specifically, HD map construction methods \cite{li2022hdmapnet,liu2023vectormapnet,MapTR,202eecvlls,20cvprpredicting,zhou2022cross,QiaoDQZ23} consider this task as the problem of predicting a collection of vectorized static map elements in bird's-eye view (BEV),  such as pedestrian crossing, lane divider, road boundaries, etc. 
On the other hand, semantic map construction methods \cite{roddick2020predicting,pan2020cross,gosala2022bird,liu2023bevfusion} treat map construction as a BEV semantic segmentation task, where each pixel in the BEV plane is assigned a semantic label. 

Based on the input sensor modality, map construction methods can be categorized into camera based~\cite{zhou2022cross,xie2022m,hao2024mapdistill}, LiDAR based~\cite{lang2019pointpillars,yin2021center} and camera-LiDAR fusion~\cite{li2022hdmapnet,liu2023vectormapnet,MapTR,2020cvprpointpainting,21nipsmultimodal,liu2023bevfusion} methods.
Camera sensors capture rich semantic information, but methods relying solely on them often struggle with spatial distortions when projecting Perspective View (PV) features into Bird's Eye View (BEV) using geometric priors. In contrast, LiDAR provides explicit geometric data with point-wise depth information, though it faces challenges related to data sparsity and sensing noise. To maximize the advantages of both modalities, recent advancements in camera-LiDAR BEV feature fusion have gained traction, effectively leveraging the semantic richness of camera data alongside the precise geometric information from LiDAR.

\begin{figure}[t]
	\setlength{\abovecaptionskip}{-0.1em}
	\begin{center}
		\includegraphics[width=0.9\linewidth]{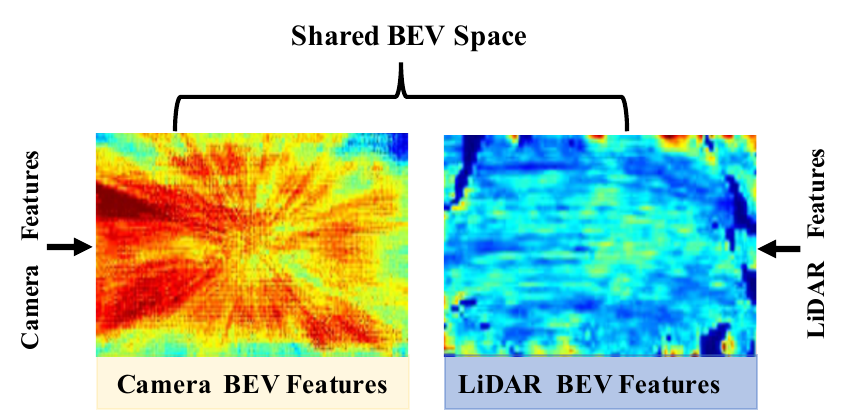}
	\end{center}
 \vspace{-2mm}
	\caption{
\textbf{Illustration of different modalities BEV features.}
Although both LiDAR and camera BEV features are presented in the shared BEV space, they may still be semantically misaligned due to the significant modality gap. (Best viewed in color. Blue color means small values and red means large.) 
	}
	\label{fig2}
\vspace{-0.5em}
\end{figure}

Recently, BEV-level fusion methods have gained significant attention for their ability to encode raw inputs from camera and LiDAR sensors into features within the same BEV space using two independent streams. 
These methods are popular because they harmonize information from different modalities while maintaining spatial consistency. 
However, as illustrated in Fig.~\ref{fig2}, LiDAR and camera BEV features can still exhibit semantic misalignment within the shared BEV space due to the substantial modality gap. 
Furthermore, existing BEV-level fusion approaches often overlook modality interaction, relying on simple element-wise operations to combine modalities, such as summation~\cite{borse2023x}, weighted averaging~\cite{bevfusion22mips}, or concatenation~\cite{liu2023bevfusion,li2022hdmapnet}. These naive fusion strategies fail to effectively address modality misalignment and do not mitigate information loss during the fusion process. 
Addressing these challenges is the motivation behind our work.

To effectively mitigate modality misalignment and information loss, a multi-modal fusion method should incorporate the following characteristics. First, it should enable interaction and integration across multiple modalities. Modality interaction involves enhancing features from one modality using information from another, thereby reducing misalignment. In contrast, modality integration fuses the well-aligned features from different modalities to produce the final output. Second, the method should employ a variety of operations, such as attention for global information exchange, convolution for effective local information aggregation, and weighting across both spatial and channel domains. This combination allows for the accumulation of each operation's strengths, leading to high-quality fusion. Currently, existing approaches only incorporate some of these elements, resulting in suboptimal fusion performance.
To address these issues, we propose a novel multi-modal BEV feature fusion method for map construction, named \textit{MapFusion}, which consists of the CIT and DDF modules to include both modality interaction and integration.
To tackle the semantic misalignment between camera and LiDAR BEV features, we propose the new Cross-modal Interaction Transform (CIT) module, which facilitates interaction between the two BEV feature spaces and enhances feature representation using a self-attention mechanism. 
Specifically, we utilize a correlation matrix to weight each position in the input multi-modal BEV features. 
This allows the CIT module to perform simultaneous intra-modality and inter-modality fusion across spatial locations, effectively capturing complementary information across different BEV modalities and mitigating modality misalignment.
To further refine the feature fusion from different modalities, we propose an effective Dual Dynamic Fusion (DDF) module to adaptively select valuable information from different modalities in a soft manner. In summary, CIT acts as modality interaction, providing flexibility by allowing fusion across both spatial locations and modalities, while DDF refines and fuses the CIT results by concentrating on modality-specific information. Both modules are indispensable for achieving optimal performance.
Importantly, the core components of MapFusion, \textit{i.e.}, CIT module and DDF module, are simple yet effective plug-and-play techniques compatible with existing pipelines for various map tasks. 
Extensive experiments on several benchmarks demonstrate the superiority of our method.

Our main contributions are summarized as follows:
\begin{itemize}

\item 
To address the Bird's-Eye View (BEV) feature fusion challenge in the multi-modal map construction task, we introduce \textit{MapFusion}, a novel method that leverages complementary information from BEV features across different modalities with both modality interaction and integration.

\item 
To solve the semantic misalignment problem between camera and LiDAR BEV features, we propose the Cross-modal Interaction Transform~(CIT) module, facilitating interaction between the two BEV feature spaces and enhancing feature representation through a self-attention mechanism.

\item 
For better feature fusion, we propose an effective Dual Dynamic Fusion (DDF) module to adaptively select valuable information from different modalities.
\item 
Compared with the state-of-the-art methods, MapFusion achieves $3.6$\% and $6.2$\% absolute improvements on the HD map construction and BEV map segmentation tasks on the nuScenes dataset, respectively, demonstrating the superiority of our approach.

\end{itemize}

The rest of this paper is organized as follows.
We briefly review related works in Section 2. 
In Section 3, we introduce our proposed method. 
We then present a variety of experimental results and analyses in Section 4.
Finally, Section 5 concludes this paper.
This paper is an extension of our preliminary work~\cite{hao2024mbfusion} published on ICRA 2024. 
The main differences between this paper and the conference version are:
\textbf{(1)\textit{ General Algorithm for BEV-based Multi-Modal Map Construction.}} More BEV-based Multi-Modal Map Construction task is realized to validate that our MapFusion, namely the CIT and DDF modules, are effective plug-and-play techniques compatible with existing pipelines for various map tasks. In the conference version we only experimented with vectorized HD map construction task, and in this paper we further include BEV map segmentation task as the new evidence of the versatility and effectiveness of our method.
\textbf{(2) \textit{Enhanced Insights into the CIT Module.}} We provide an internal diagram (see Fig.~\ref{cit-frame}) and equation (see Eq.~\ref{cit-a}) to clarify the theoretical foundations of the CIT module, enhancing understanding of its mechanisms.
The main idea behind our CIT module is to leverage the self-attention mechanism to learn the binary relationships between camera and LiDAR modalities. 
Specifically, we utilize a correlation matrix to weight each position of the input feature maps, formulated as Eq.~\ref{cit-a}, where $\alpha_{i,j}$ represents the correlation between the $i$-th and $j$-th positions on the feature maps. 
This leads to the inference of four matrix blocks when calculating the correlation matrix $\alpha$: two intra-modality correlation matrix blocks (for Camera and LiDAR) and two inter-modality correlation matrix blocks, as illustrated in Fig.~\ref{cit-frame}. 
Consequently, the CIT module can adaptively perform simultaneous intra-modality and inter-modality information fusion, comprehensively capturing complementary information between BEV features of different modalities.
\textbf{(3)	\textit{Extensive Ablation Studies and Analysis.}} We conduct additional ablation studies to validate the effectiveness of each proposed component across two BEV-based multi-modal map construction tasks. These studies include: the contributions of CIT and DDF (See Tab.~\ref{tab5} and Tab.~\ref{tab6}), variations of different fusion methods (See Tab.~\ref{tab7} and Tab.~\ref{tab8}), compatibility with other HD map construction methods (See Tab.~\ref{tab9}), and an analysis of the accuracy-computation trade-off using our proposed CIT module and different fusion strategies (See Fig.~\ref{fig5}). Based on these ablation experiments, we also conduct a deeper analysis of the working mechanism of our method.
\textbf{(4)	\textit{More Visualization Results.}}  We include additional visualization results to further illustrate our findings. Fig.~\ref{tsne} shows the visualization results of the t-SNE and the feature maps before and after the CIT module, demonstrating the CIT module's ability to mitigate the misalignment between different modalities.
Fig.~\ref{vis} illustrates the feature maps before and after the CIT module, which integrates various modes of BEV features into a unified space.
In addition, we present the qualitative results of the CIT and DDF modules for the BEV map segmentation and HD map tasks in Fig.~\ref{fig6} and Fig.~\ref{fig7}, respectively.

\begin{figure*}[t]
	\begin{center}
		\includegraphics[width=1.0\linewidth]{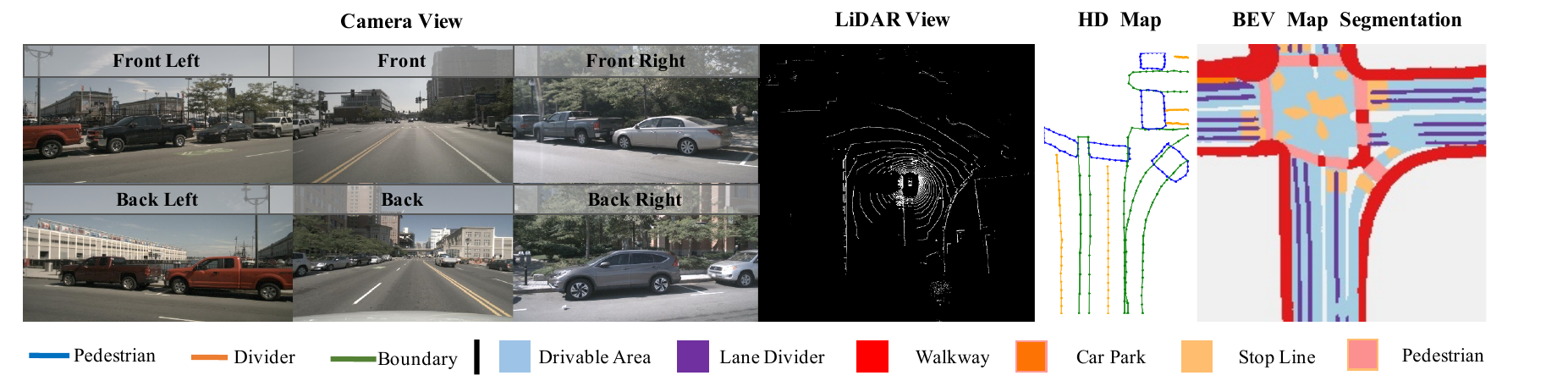}
	\end{center}
	\vspace{-4mm}
	\caption{
Illustration of different map construction tasks (HD map construction and BEV map segmentation). 
	}
	\label{fig1}
\end{figure*}

\section{Related Work}
\label{sec2}
Our work is highly related to map construction task (See  Fig.~\ref{fig1}) and multi-sensor fusion methods, which will be discussed thoroughly in the following.

\subsection{Map Construction Task}
\textbf{HD map construction.}
HD map construction is a critical and extensively researched area in autonomous driving.
Based on input sensor modalities,
HD map construction models can be categorized into camera-based~\cite{zhang2023online,ding2023pivotnet,qiao2023end,maptrv2,yuan2024streammapnet}, LiDAR-based~\cite{lang2019pointpillars,yin2021center} and camera-LiDAR fusion~\cite{li2022hdmapnet,liu2023vectormapnet,MapTR,hao2025msc,hao2024your} models.
Camera-only methods~\cite{zhang2023online,ding2023pivotnet,qiao2023end,maptrv2,yuan2024streammapnet} have increasingly adopted the Bird's-eye view (BEV) representation as an ideal feature space for multi-view perception, owing to its remarkable ability to mitigate scale ambiguity and occlusion challenges. 
Various techniques have been proposed to project perspective view (PV) features onto the BEV space by leveraging geometric priors, such as LSS~\cite{philion2020lift}, Deformable Attention~\cite{22eccvbevformer}, and GKT~\cite{2022GKT}.
Nevertheless, camera-only methods lack explicit depth information, which forces them to rely on higher resolution images or larger backbone models to achieve enhanced accuracy~\cite{liu2021Swin,liu2021swinv2,hao2023mixgen,wang2022internimage,22eccvbevformer,yang2022bevformer,xiong2023neural}. 
In contrast, LiDAR-only approaches~\cite{lang2019pointpillars,yin2021center} benefit from the accurate 3D geometric information provided by LiDAR input.
However, they face challenges related to data sparsity and sensing noise.

Recently, camera-LiDAR fusion methods~\cite{li2022hdmapnet,liu2023vectormapnet,MapTR} leverage the semantic richness of camera data and the geometric information from LiDAR in a collaborative manner. 
BEV-level fusion, which uses two independent streams to encode raw inputs from camera and LiDAR sensors into features within the same BEV space, has gained significant attention~\cite{liu2023bevfusion,bevfusion22mips}.
This approach incorporates complementary modality features, outperforming uni-modal input approaches.
Existing HD map construction multi-sensor fusion methods—HDMapNet~\cite{li2022hdmapnet}, VectorMapNet~\cite{liu2023vectormapnet}, and MapTR~\cite{MapTR}— utilize straightforward channel concatenation and convolution for multi-modal feature fusion. 
However, these methods overlook modality interaction and employ very simple fusion strategies, leading to issues of misalignment and information loss.

\textbf{BEV map segmentation.}
Semantic map construction methods \cite{roddick2020predicting,pan2020cross,gosala2022bird,liu2023bevfusion} take map construction as a BEV semantic segmentation task, assigning semantic labels to each pixel in the BEV plane.
Building on Perspective View (PV) segmentation~\cite{roddick2018orthographic,202eecvlls}, early approaches utilize homography transformations to convert camera images into bird's-eye view (BEV) representations, followed by the estimation of segmentation maps~\cite{ammar2019geometric,zhang2022beverse,garnett20193d,zhu2021monocular}.
However, homography transformation introduces strong artifacts, and BEV-based methods~\cite{zhou2022cross, xie2022m,liu2023bevfusion}, i.e. performing segmentation directly on BEV plane, have received extensive attention.
CVT~\cite{zhou2022cross} employs a learned map embedding and an attention mechanism between map queries and camera features.
Furthermore, BEVFusion~\cite{liu2023bevfusion}, BEVerse~\cite{zhang2022beverse} and M$^{2}$BEV~\cite{xie2022m} explore
multi-task learning with 3D object detection.
However, these approaches lack explicit utilization of depth information, resulting in unsatisfactory performance.

Existing fusion methods~\cite{2020cvprpointpainting,21nipsmultimodal} primarily focus on object-centric and geometry-oriented approaches. For instance, PointPainting~\cite{2020cvprpointpainting} enhances only the foreground LiDAR points, while MVP~\cite{21nipsmultimodal} concentrates solely on densifying foreground 3D objects. 
Both methods also assume that LiDAR is the more effective modality for sensor fusion, which may not be valid for map construction tasks~\cite{liu2023bevfusion}.
Additionally, X-Align~\cite{borse2023x} employs an integration method that combines the features of the two modalities before applying attention, neglecting modality interactions and relying on overly simplistic fusion strategies.
In summary, these methods utilize basic feature concatenation to merge multi-modal features, necessitating the network to implicitly reconcile information from misaligned features.

\subsection{Multi-sensor Fusion}
Multi-sensor fusion has garnered significant attention in the field of autonomous driving. Existing approaches can be broadly categorized into three types: point-level fusion, feature-level fusion, and BEV-level fusion.
Point-level fusion methods~\cite{21cvprpointaug,2020cvprpointpainting,21itscfusionpainting,22ijcaiautoalign,21nipsmultimodal} typically project image semantic features onto foreground LiDAR points, enabling LiDAR-based detection on the enhanced point cloud. While effective for 3D object detection tasks, these methods are less suitable for semantically driven tasks such as BEV map segmentation~\cite{liu2023bevfusion,bevfusion22mips,zhou2022cross,xie2022m} and HD map construction~\cite{li2022hdmapnet,liu2023vectormapnet,MapTR}. This limitation stems from the lossy projection of camera features to LiDAR, where only about 5\% of camera features align with points from a typical 32-beam LiDAR scanner, resulting in significant information loss.
Feature-level fusion methods~\cite{22cvprfocal,18eccvdeepcontinuous} first project LiDAR points into a feature space or generate proposals, query the corresponding camera features, and then concatenate them back into the feature space. However, both point-level and feature-level fusion approaches encounter generalization challenges. Specifically, point-level fusion is not easily extendable to other sensor modalities, while feature-level fusion struggles with generalization across different tasks.

Recently, multi-modal feature fusion in a unified BEV space has gained considerable attention~\cite{MapTR,liu2023vectormapnet,li2022hdmapnet,liu2023bevfusion,bevfusion22mips}. BEV-level fusion employs two independent streams to encode raw inputs from camera and LiDAR sensors into features within the same BEV space. This approach offers a straightforward yet effective means to integrate BEV-level features from both streams, facilitating their use in various downstream tasks.
However, existing BEV-level fusion methods often overlook modality interactions, relying on element-wise operations (such as summation or mean) or simple concatenation. This can lead to issues of misalignment and information loss. In this paper, we propose a simple and effective camera-LiDAR BEV feature fusion method that simultaneously integrates complementary information from different modalities, specifically targeting multi-modal map construction tasks.

\noindent\textbf{Comparison with Existing Works.}
This work differs from prior literature in \textit{three} key aspects. 
Firstly, we focus on the BEV-based multi-modal map construction task, distinct from other BEV perception tasks~\cite{chen2025stvit+,21cvprpointaug,21itscfusionpainting}, as it aims at predicting map elements, such as pedestrian crossing, lane divider, road boundaries, etc. 
In fact, the map construction task is a semantic-oriented task, which pays more attention to the semantic information in the image.
Therefore, the performance of directly using the fusion method on the 3D object detection task to the map task is not satisfactory.
Secondly, to solve the semantic misalignment problem between Camera and LiDAR BEV features, we propose Cross-modal Interaction Transform (CIT) module to enable the two BEV feature spaces to interact with each other and enhance feature representation through a self-attention mechanism.
Additionally, to further fuse features from different modalities, we propose an effective Dual Dynamic Fusion (DDF) module to adaptively select valuable information from different modalities.
To the best of our knowledge, \textit{MapFusion} is the first to explore the effectiveness of interactive modules on multi-modal map construction tasks.
Last but not least, the core components of \textit{MapFusion}, \textit{i.e.}, CIT module and DDF module, are simple yet effective plug-and-play techniques compatible with existing pipelines for various map tasks, such as HD map and semantic map construction.

\section{Methodology}
\label{sec3}
\begin{figure*}[t]
	\centering
 \vspace{2em}
 	\includegraphics[width=1\textwidth]{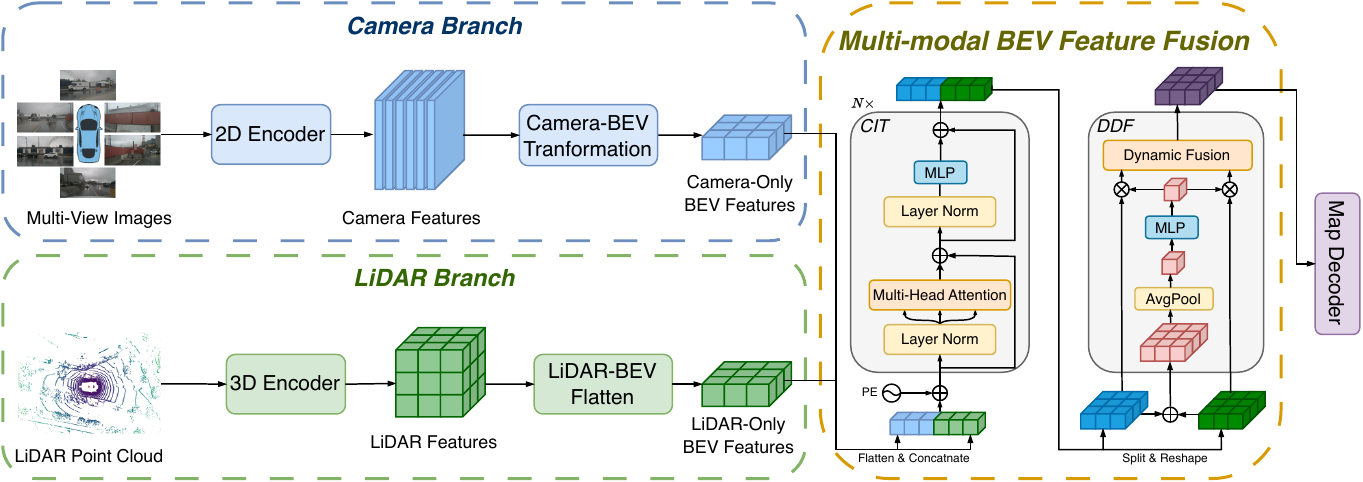}
	\caption{
\textbf{An overview of MapFusion framework.} 
First, we extract features from multi-modal inputs and convert them into a shared bird's-eye view (BEV) space efficiently using view transformations.  
To fuse the BEV features from different modalities, we first propose Cross-modal Interaction Transform (CIT) module to enhance one modality from another modality by self-attention mechanism.
Afterwards, we propose a Dual Dynamic Fusion (DDF) module to 
automatically select valuable information from different modalities for better feature fusion.
Finally, the fused multi-modal BEV features are fed into a shared decoder and prediction heads for map construction tasks. 
}
	\label{fig3}
\end{figure*}

We propose a novel multi-modal BEV map construction approach called \textit{MapFusion}, which is a simple yet effective plug-and-play technique compatible with existing pipelines for various map construction tasks. 
The overview framework of \textit{MapFusion} is shown in Fig.~\ref{fig3}. 
Given different sensory inputs, we first apply modality-specific encoders to extract their features. 
These multi-modal features are then transformed into a unified BEV representation that preserves both geometric and semantic information. 
Then, we propose Cross-modal Interaction
Transform (CIT) module to make these two BEV feature spaces exchange knowledge with each other to enhance the feature representation by the self-attention mechanism.
Additionally, we introduce a novel Dual Dynamic Fusion (DDF) module to automatically select valuable information from different modalities, which can take full advantage of the inherent complementary information between different modalities.
Finally, the fused multi-modal BEV features are fed into decoder and prediction heads for map construction tasks.

\subsection{Preliminaries}
\label{3.1}
For notation clarity, we first introduce some symbols and definitions used throughout this paper.  
Our goal is to design a novel framework taking multi-modal
sensor data $\chi$ as input and predicting map elements in BEV space, and the types of the map elements (supported types are road boundary, lane divider, and pedestrian crossing, etc).
Formally, assume that we have a set of inputs, $\chi = \{\textit{Camera}, \textit{LiDAR}\}$, containing multi-view RGB camera images in perspective view, $\textit{Camera} \in \mathbb{R}^{N^{\mathrm{cam}}\times H^{\mathrm{cam}}\times W^{\mathrm{cam}}\times 3}$, $N^{\mathrm{cam}}$, $H^{\mathrm{cam}}$, $W^{\mathrm{cam}}$ denote number of cameras, image height, and image width, respectively, as well as a LiDAR point cloud,
$\textit{LiDAR} \in \mathbb{R}^{P\times 5}$, with number of points $P$. 
Each point consists of its 3-dimensional coordinates, reflectivity, and beam index. The detailed architectural designs are described as follows.

\subsection{Map Encoder}\label{sec3.2}
We apply modality-specific encoders to extract their features and transform multi-modal features into a unified BEV representation that preserves both geometric and semantic information. Note that our approach is compatible with other Map Encoders that can also be employed to generate camera-only and LiDAR-only BEV features.

\textbf{Camera to BEV.}
We extract BEV features from multi-view RGB images with the BEV feature extractor.
It consists of a backbone~\cite{he2016deep,liu2021Swin} to extract multi-scale
2D features from each perspective view, an FPN~\cite{lin2017feature} to
refine and fuse multi-scale features into single-scale features,
and a 2D-to-BEV feature transformation module~\cite{202eecvlls,2022GKT} to map 2D features into BEV features.
The camera BEV features can be denoted as $\mathbf{F}_{\mathrm{Camera}}^{\mathrm{BEV}} \in \mathbb{R}^{ H\times W\times C}$, where
$H, W, C$ refer to the spatial height, spatial width, and the
number of channels of BEV feature maps, respectively.

\textbf{LiDAR to BEV.}
For the LiDAR points, we follow SECOND~\cite{2018seoncd} in using voxelization and a sparse LiDAR encoder. 
The LiDAR features are projected to BEV space using a flattening operation as in \cite{liu2023bevfusion}, to obtain the unified LiDAR BEV representation $\mathbf{F}_{\mathrm{LiDAR}}^{\mathrm{BEV}} \in \mathbb{R}^{H\times W\times C}$.

\subsection{ Cross-modal Interaction Transform (CIT)}\label{sec3.3}
Existing methods directly convert all sensory features to the shared BEV representation, and then fuse them via arithmetic or splicing operations to obtain multi-modal BEV features. 
However, despite being in the same BEV space, LiDAR BEV features and camera BEV features can still be semantically misaligned due to the significant modality gap, leading to a misalignment problem.
To address this issue, we propose a new and powerful Cross-Modal
Interaction Transformer (CIT) module to enhance one modality from another modality by the self-attention mechanism. 
Next, we describe in detail our proposed CIT module.

\textbf{Concatenation Interaction Transformer.} First, given the BEV features from both camera (
$\mathbf{F}_{\mathrm{Camera}}^{\mathrm{BEV}} \in \mathbb{R}^{H\times W\times C}$
) and LiDAR
(
$\mathbf{F}_{\mathrm{LiDAR}}^{\mathrm{BEV}} \in \mathbb{R}^{H\times W\times C}$
) sensors,
the BEV tokens 
$\mathbf{T}_{\mathrm{Camera}}^{\mathrm{BEV}} \in \mathbb{R}^{HW\times C}$
and
$\mathbf{T}_{\mathrm{LiDAR}}^{\mathrm{BEV}} \in \mathbb{R}^{HW\times C}$
are obtained by flattening each BEV feature and permuting the order of the matrices. 
Second, we concatenate the tokens of each modality and add a learnable positional embedding, which is a trainable parameter of dimension $2HW\times C$, to get the input BEV tokens 
$\mathbf{T}^{\mathrm{in}} \in \mathbb{R}^{2HW\times C}$
of the Transformer~\cite{vaswani2017attention}.
The positional embedding enables the model to differentiate spatial information between different tokens at training time. 
Third, the input token 
$\mathbf{T}^{\mathrm{in}}$
uses linear projections for computing a set of queries, keys and values (
$\mathbf{Q}$, $\mathbf{K}$ and $\mathbf{V}$
),
\begin{equation} \label{eq1}
\mathbf{Q} = \mathbf{T}^{\mathrm{in}}\mathbf{W}^\mathrm{Q},
\mathbf{K} = \mathbf{T}^{\mathrm{in}}\mathbf{W}^\mathrm{K},
\mathbf{V} = \mathbf{T}^{\mathrm{in}}\mathbf{W}^\mathrm{V},
\end{equation}
where 
$\mathbf{W}^\mathrm{Q} \in \mathbb{R}^{C\times D_\mathrm{Q}}$, $\mathbf{W}^\mathrm{K} \in \mathbb{R}^{C\times D_\mathrm{K}}$ and  $\mathbf{W}^\mathrm{V} \in \mathbb{R}^{C\times D_\mathrm{V}}$
are weight
matrices.
Moreover, 
$D_\mathrm{Q}$, $D_\mathrm{K}$ and $D_\mathrm{V}$
are equal in our Transformer, i.e., 
$D_\mathrm{Q} = D_\mathrm{K} = D_\mathrm{V} = C$.
Fourth, the self-attention layer uses the scaled dot products between 
$\mathbf{Q}$ and $\mathbf{K}$
to compute the attention weights and then multiply by the values to infer
the refined output 
$\mathbf{Z}$,
\begin{equation} \label{eq2}
\mathbf{Z} = \operatorname{Attention}(\mathbf{Q}, \mathbf{K}, \mathbf{V}) = \operatorname{softmax}\left(\frac{\mathbf{Q}\mathbf{K}^T}{\sqrt{D_\mathrm{k}}}\right)\mathbf{V},
\end{equation}
where 
$\frac{1}{\sqrt{D_\mathrm{k}}}$
is a scaling factor for preventing the softmax function from falling into a region with extremely small gradients when the magnitude of dot products grows large. To encapsulate multiple complex relationships from different representation subspaces at different positions, the multi-head attention mechanism is adopted,

\begin{equation} \label{eq3}
\begin{aligned}
  \hat{\mathbf{Z}} &= \operatorname{MultiHead}(\mathbf{Q}, \mathbf{K}, \mathbf{V}) = \operatorname{Concat}(\mathbf{Z}_1, \cdots, \mathbf{Z}_h)\mathbf{W}^\mathrm{O},\\
  \mathbf{Z}_i &= \operatorname{Attention}(\mathbf{Q}\mathbf{W}_i^\mathrm{Q}, \mathbf{K}\mathbf{W}_i^\mathrm{K}, \mathbf{V}\mathbf{W}_i^\mathrm{V}), i \in \{1, \cdots, h\}.
\end{aligned}
\end{equation}

The subscript $h$ denotes the number of heads, and
$\mathbf{W}^\mathrm{O} \in \mathbb{R}^{h \cdot C \times C}$
denotes the projected matrix of 
$\operatorname{Concat}(\mathbf{Z}_1, \cdots, \mathbf{Z}_h)$.
Finally, the transformer uses a non-linear transformation to
calculate the output features, 
$\mathbf{T}^{\mathrm{out}}$ 
which are of the same shape as the input features 
$\mathbf{T}^{\mathrm{in}}$,
\begin{equation} \label{eq4}
\mathbf{T}^{\mathrm{out}} = \operatorname{MLP}(\hat{\mathbf{Z}}) + \mathbf{T}^{\mathrm{in}}.
\end{equation}
The output 
$\mathbf{T}^{\mathrm{out}}$ 
are converted into 
$\hat{\mathbf{F}}_{\mathrm{Camera}}^{\mathrm{BEV}}$ and $\hat{\mathbf{F}}_{\mathrm{LiDAR}}^{\mathrm{BEV}}$
for further feature fusion.

\begin{figure}[t]
	\centering
 	\includegraphics[width=0.48\textwidth]{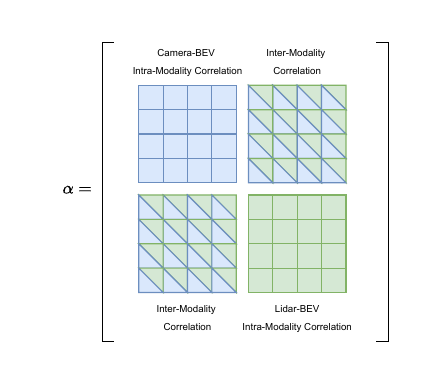}
	\caption{
Illustration of the Correlation Matrix $\alpha$.
}
	\label{cit-frame}
\end{figure}

\begin{figure}[h]
 \color{black}
 \NiceMatrixOptions{code-for-first-row = \color{red},
  code-for-first-col = \color{blue}}
 \begin{equation}     \label{cit-a}
  \boldsymbol{\alpha}  = \operatorname{softmax}\left(\frac{\mathbf{Q K}^{T}}{\sqrt{D_K}}\right)  =
  \begin{pNiceArray}{cccc|cccc}[first-row,first-col,nullify-dots]
   & C^{BEV}_1 &  &  \Cdots & C^{BEV}_{HW} &  C^{BEV}_{HW+1} &   & \Cdots &   C^{BEV}_{2HW}  \\
   L^{BEV}_1 & \alpha_{1,1} & \alpha_{1,2} & \cdots & \alpha_{1,HW} &   \alpha_{1,HW+1}  & \alpha_{1,HW+2}& \cdots & \alpha_{1,2HW}  \\
   & \alpha_{2,1} & \alpha_{2,2} & \cdots & \alpha_{2,HW} &   \alpha_{2,HW+1}  & \alpha_{2,HW+2}& \cdots & \alpha_{2,2HW}  \\
   \Vdots & \vdots & \vdots & \ddots & \vdots &  \vdots  & \vdots & \ddots & \vdots  \\
   L^{BEV}_{HW} & \alpha_{HW,1} & \alpha_{HW,2} & \cdots & \alpha_{HW,HW} &   \alpha_{HW,HW+1}  & \alpha_{HW,HW+2}& \cdots & \alpha_{HW,2HW}  \\
   \hline
   L^{BEV}_{HW+1} & \alpha_{HW+1,1} & \alpha_{HW+1,2} & \cdots & \alpha_{HW+1,HW} &   \alpha_{HW+1,HW+1}  & \alpha_{HW+1,HW+2}& \cdots & \alpha_{HW+1,2HW}  \\
   & \alpha_{HW+2,1} & \alpha_{HW+2,2} & \cdots & \alpha_{HW+2,HW} &   \alpha_{HW+2,HW+1}  & \alpha_{HW+2,HW+2}& \cdots & \alpha_{HW+2,2HW}  \\
   \Vdots & \vdots & \vdots & \ddots & \vdots &  \vdots  & \vdots & \ddots & \vdots  \\
   L^{BEV}_{2HW}  & \alpha_{2HW,1} & \alpha_{2HW,2} & \cdots & \alpha_{2HW,HW} &   \alpha_{2HW,HW+1}  & \alpha_{2HW,HW+2}& \cdots & \alpha_{2HW,2HW}  \\
  \end{pNiceArray},
 \end{equation}
\end{figure} 

\textit{Remarks}:
The main idea behind our CIT module is to leverage the self-attention mechanism to learn the binary relationships between Camera and LiDAR modalities. More specifically, we utilize a correlation matrix to weight each position in the input feature maps. This can be formulated as shown in Eq.~\ref{cit-a}. In this formula, $\alpha_{i,j}$ represents the correlation between the $i$-th position and the $j$-th position on the feature maps. 
According to Eq.~\ref{cit-a}, four matrix blocks can be naturally inferred when calculating the correlation matrix $\alpha$. Two of these blocks represent intra-modality correlation matrices (for Camera and LiDAR), while the other two represent inter-modality correlation matrices, as illustrated in Fig.~\ref{cit-frame}. 
Thus, we utilize the correlation matrix to weight each position of the input multi-modal BEV features. The CIT module can then adaptively perform simultaneous intra-modality and inter-modality information fusion, robustly capturing the complementary information between BEV features of different modalities.

\subsection{ Dual Dynamic Fusion (DDF)}\label{sec3.4}
\begin{figure*}[t]
	\centering
 	\includegraphics[width=0.96\textwidth]{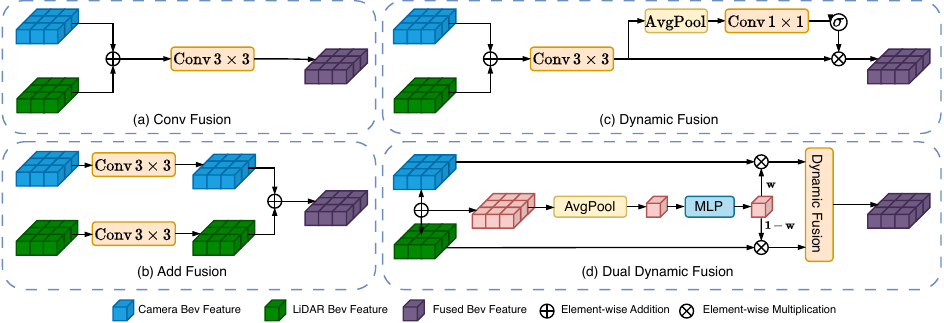}
	\caption{
Three existing fusion strategies and our proposed Dual Dynamic Fusion (DDF) strategy.
}
	\label{fig4}
\end{figure*}

Despite the effectiveness of the cross-modal interaction transform module, we argue that how to design an effective cross-modal fusion strategy to adaptively select valuable information from different modalities for better feature fusion is still very important.
Recently, multi-modal BEV feature fusion methods~\cite{bevfusion22mips,liu2023bevfusion} have received much attention. 
It is a common approach to utilize concatenation followed by convolution to combine features from multi-modal BEV feature inputs, $\hat{\mathbf{F}}_{\mathrm{Camera}}^{\mathrm{BEV}}$ and $\hat{\mathbf{F}}_{\mathrm{LiDAR}}^{\mathrm{BEV}}$, resulting in the aggregated features $\mathbf{F}_{\mathrm{fused}}$, as shown in Fig.~\ref{fig4} (a) Conv Fusion.
Another common method is to use CNN to convolve the BEV features of different modalities separately, and then add the convolutional features, as shown in Fig.~\ref{fig4} (b) Add Fusion.
As Fig. ~\ref{fig4} (c) illustrates, 
the input of the Dynamic Fusion (DF) module is the Conv Fusion output features, and then they are fused with learnable static weights, inspired by Squeeze-and-Excitation mechanism~\cite{hu2018squeeze}.
To effectively select valuable information from different modalities, we propose a Dual Dynamic Fusion (DDF) module for better feature fusion and maximum performance gain. 
Next, we describe in detail our proposed fusion designs.

\textbf{Dual Dynamic Fusion.}
As shown in Fig.~\ref{fig4}(d), our Dual Dynamic Fusion (DDF) module takes two sets of features from the camera BEV features and  LiDAR BEV features as input. 
In order to generate meaningful attention weights that can effectively select informative features from both inputs, we first sum the features from both branches before performing the squeeze and excitation operations that generate the attention weights. 
We can formulate this process as:
\begin{equation}
\label{eq11}
\mathbf{w} = \sigma\left(\gamma\left(\operatorname{AvgPool}\left( \hat{\mathbf{F}}_{\mathrm{Camera}}^{\mathrm{BEV}} +\hat{\mathbf{F}}_{\mathrm{LiDAR}}^{\mathrm{BEV}} \right)\right)\right),
\end{equation}
where $\sigma$ and $\gamma$ represent the sigmoid function and linear layers respectively, $\operatorname{AvgPool}$ is the global average pooling operation, and $\mathbf{w}$ denotes the attention weights. 
We then multiply $\mathbf{w}$ and $\mathbf{1}-\mathbf{w}$ to both input features before the summation so that the fusion process essentially acts as a self-gating
mechanism to adaptively select useful information from different
BEV features:
\begin{equation}
\label{eq12}
\mathbf{F}_{\mathrm{fused}} = \operatorname{Adaptive}\left(\operatorname{Conv_{3\times3}}\left(\left[\mathbf{w}\cdot\hat{\mathbf{F}}_{\mathrm{Camera}}^{\mathrm{BEV}},(\mathbf{1}-\mathbf{w})\cdot\hat{\mathbf{F}}_{\mathrm{LiDAR}}^{\mathrm{BEV}}\right]\right)\right),
\end{equation}
where $[\cdot, \cdot]$ denotes the concatenation operation along the channel dimension. $\cdot$ is element-wise multiplication. 
$\operatorname{Conv_{3\times3}}$ fuses the channel and spatial information with a $3 \times 3$ convolution layer to reduce the channel dimension of concatenated feature to $C$. With input feature 
$\hat{\mathbf{F}} \in \mathbb{R}^{H \times W\times C}$, the $\operatorname{Adaptive}$ operation is formulated as:
\begin{equation}
\label{eq13}
\operatorname{Adaptive}(\hat{\mathbf{F}}) =\sigma\left(\mathbf{W}\operatorname{AvgPool}(\hat{\mathbf{F}})\right)\cdot \hat{\mathbf{F}},
\end{equation}
where $\mathbf{W}$ denotes linear transform matrix (e.g., $1 \times 1$ convolution) and $\sigma$ denotes sigmoid function.
Therefore, the DDF module can adaptively select valuable information from two modalities for better feature fusion. 
The output fused feature $\mathbf{F}_{\mathrm{fused}}$ will be used for map construction task, with the decoder and prediction heads.

\textit{Remarks}:
DF only performs channel-wise fusion, while DDF first conducts spatial fusion and then channel-wise fusion. DDF enhances DF by incorporating global average pooling, utilizing global weights to reduce information loss. 
In DDF module, the AvgPool in Eq~\ref{eq11} is performed in the spatial domain with an input dimension of $W \times H \times C$ and an output of $C$; The AvgPool in Eq.~\ref{eq13} is performed in the channel domain with an input of $W \times H \times C$ and an output of $W \times H $.

\subsection{Map-Task Heads}\label{sec3.5}
We apply specific heads for different map tasks to the fused BEV features.
We show two examples:  HD map construction and BEV map segmentation.

\textbf{HD map construction head.}
HD map constructors formulate this task as predicting a collection of vectorized static map elements in bird's eye view (BEV), \textit{i.e.}, pedestrian crossings, lane dividers, road boundaries.
We follow MapTR~\cite{MapTR} to train the map head with the classification loss~\cite{mukhoti2020calibrating}, the point2point loss~\cite{malkauthekar2013analysis}, and the edge direction loss~\cite{MapTR}.

\textbf{BEV map segmentation head.}
Different map categories may overlap (e.g., crosswalk is a subset of drivable space).
Therefore, we formulate this problem as multiple binary semantic segmentation, one for each class.
We follow BEVFusion~\cite{liu2023bevfusion} to train the segmentation head with the standard focal loss~\cite{mukhoti2020calibrating}.

\section{Experiments}
\subsection{Dataset}
\textbf{NuScenes Datasets.}
We evaluate our method on the widely-used challenging nuScenes~\cite{20cvprnuscense} dataset following the standard settings of previous methods~\cite{liu2023bevfusion,MapTR}. 
The nuScenes dataset contains 1,000 sequences of recordings collected by autonomous driving cars. 
Each sample is annotated at 2Hz and contains 6 camera images covering $360^\circ$ horizontal FOV of the ego-vehicle.
For the HD map construction task, we following MapTR~\cite{MapTR} and three kinds of map elements are chosen for fair evaluation – pedestrian crossing, lane divider, and road boundary. 
Moreover, for the BEV map segmentation task, we following BEVFusion~\cite{liu2023bevfusion}, we predict six semantic classes: drivable lanes, pedestrian crossings, walkways, stop lines, carparks, and lane dividers. 

\textbf{Argoverse2 Dataset.} 
There are 1,000 logs in the Argoverse2 dataset~\cite{21nipsarfoverse}.
Each log contains 15s of 20Hz RGB images from 7 cameras, 10Hz LiDAR sweeps, and a 3D vectorized map. 
The train, validation, and test sets contain 700, 150, and 150 logs, respectively. 
For both HD map construction and BEV map segmentation tasks, we select three map elements for fair evaluation: pedestrian crossing, lane divider, and road boundary.

\subsection{Evaluation Metrics}
\textbf{HD map construction task.} 
We adopt the evaluation metrics consistent with previous works~\cite{li2022hdmapnet,liu2023vectormapnet,MapTR}, where average precision (AP) is used to evaluate the map construction quality and Chamfer distance $D_{\textrm{Chamfer}}$ determines the matching between predictions and ground truth. We calculate the $\textrm{AP}_\tau$ under
several $D_{\textrm{Chamfer}}$ thresholds ($\tau \in T =\{0.5\textrm{m},1.0\textrm{m},1.5\textrm{m}\} $), and then average across all thresholds as the final mean AP (\textit{mAP}) metric,
\begin{equation}
\label{eq19}
\begin{array}{l}
\textit{mAP}\ =\ \frac{1}{\left|T\right|}\sum\limits_{\tau\in T}{\textrm{AP}}_\tau.
\end{array}
\end{equation}
The perception ranges are $[-15.0\mathrm{m}, 15.0\mathrm{m}]/[-30.0\mathrm{m}, 30.0\mathrm{m}]$ for X/Y-axes.

\textbf{BEV map segmentation task.} 
For the BEV map segmentation task, our primary evaluation metric is the mean Intersection over Union (mIoU). 
Due to potential overlaps between classes, we apply binary segmentation separately to each class and choose the highest IoU over different thresholds.
We then average these values over all semantic classes to produce the mIoU. 
This evaluation protocol was proposed in BEVFusion~\cite{liu2023bevfusion}. 

\subsection{Experimental Setting}
MapFusion is trained with $4$ NVIDIA RTX A6000 GPUs. 
For the HD map construction task, we build upon MapTR~\cite{MapTR} as the baseline.
Specifically, we adopt ResNet50~\cite{2016cvprresnet} and SECOND~\cite{2018seoncd} as the backbone and employ GKT~\cite{2022GKT} as the default 2D-to-BEV module. 
Training losses include classification loss, point2point loss, and edge direction loss. 
with weights of 2.0, 5.0, and 0.005, respectively.
The model is trained for $24$ and $6$ epochs on the nuScenes and Argoverse2 datasets
respectively. 
All the data pre-processing steps for both
datasets follow MapTR~\cite{MapTR}.
We set the mini-batch size to $16$, and use a step-decayed learning rate with an initial value of $4e^{-3}$.
For the BEV map segmentation task, we use BEVFusion~\cite{liu2023bevfusion} as our baseline and train our networks within the mmdetection3d framework~\cite{contributors2020mmdetection3d}. Specifically, we adopt Swin-T~\cite{liu2021Swin} and VoxelNet~\cite{2018seoncd} as the backbone, and utilize LSS~\cite{202eecvlls} as the default 2D-to-BEV module. The model is trained for 20 and 6 epochs on the nuScenes and Argoverse2 datasets, respectively.
The baseline is trained using the hyperparameters reported in~\cite{liu2023bevfusion}, following a learning schedule of 20 epochs with a cyclic learning rate, starting for $1e^{-4}$ and performing a single cycle with target ratios {10, $1e^{-4}$} and a step of 0.4
For the CIT module described in Section~\ref{sec3.3} of the paper, we added this module before the fuser operation in the baseline model.
To implement the cross-modal interaction,  we first obtain BEV tokens by flattening each BEV feature and permuting the order of the matrix.
Then, we concatenate the tokens of each modality and add a learnable positional embedding.
This step is followed by a multi-head self-attention block as described in~\cite{vaswani2017attention}, containing 8 heads and an embedding dimension of $256$.
For the DDF module described in Section~\ref{sec3.4} of the paper, we replace the the naive convolutional fuser with the DDF module in the baseline model.

\begin{table*}
\begin{center}
\caption{
\textbf{Comparisons with state-of-the-art methods on nuScenes val set for the HD map construction task.}
We compare with existing methods from literature, where the numbers are taken from MapTR~\cite{MapTR}.
We also provide information on the backbones, epochs and input modalities in the table. 
Our proposed MapFusion outperforms all existing approaches in both single-class APs and the overall mAP by a significant margin.
}
\label{tab1}
\scalebox{0.83}{
 \begin{tabular}{r|ccc|ccc|l}
  \hline
\rowcolor{black!10}  Method &  Modality & Backbone & Epochs & AP$_{ped}$  & AP$_{divider}$ & AP$_{boundary}$& mAP\\
 \hline
  \midrule
HDMapNet~\cite{li2022hdmapnet}& C & Efficient-B0 & 30 & 14.4 & 21.7 & 33.0 & 23.0\\
HDMapNet~\cite{li2022hdmapnet}& L & PointPillars & 30 & 10.4 & 24.1 & 37.9 & 24.1\\
HDMapNet~\cite{li2022hdmapnet}& C \& L & Efficient-B0 \& PointPillars & 30 & 16.3 & 29.6 & 46.7 & 31.0\\
\hline
VectorMapNet~\cite{liu2023vectormapnet} & C & ResNet-50 & 110 & 36.1 & 47.3 & 39.3 & 40.9\\
VectorMapNet~\cite{liu2023vectormapnet}& L & PointPillars & 110 & 25.7 & 37.6 & 38.6 & 34.0\\
VectorMapNet~\cite{liu2023vectormapnet}&  C \& L & ResNet-50 \& PointPillars & 110 & 37.6  & 50.5 & 47.5 & 45.2\\
   \hline
MapTR~\cite{MapTR}&  \makecell[c]{C}& \makecell[c]{ResNet-50} & \makecell[c]{24}& \makecell[c]{46.3} &\makecell[c]{51.5}&\makecell[c]{53.1}&\makecell[l]{50.3}\\
MapTR~\cite{MapTR}& \makecell[c]{L}& \makecell[c]{SECOND} & \makecell[c]{24}& \makecell[c]{48.5} &\makecell[c]{53.7}&\makecell[c]{64.7}&\makecell[l]{55.6}\\
MapTR~\cite{MapTR}& C \& L & ResNet-50 \& SECOND  & 24 & 55.9 & 62.3 & 69.3 & 62.5\\
    \hline
\rowcolor{blue!10} \textbf{MapFusion (Ours)} &  C \& L &  ResNet-50 \& SECOND & 24 & \textbf{61.6} & \textbf{64.4} & \textbf{72.5} & \textbf{66.1}$_{+3.6}$\\
 
  \bottomrule 
  \end{tabular}}
\end{center}

\end{table*}

\begin{table*}
\begin{center}
\caption{
\textbf{Results of the HD map construction task on the Argoverse2 dataset.}
$\dag$  denotes our re-implementation following the setting in the paper.}
\label{tab2}
\scalebox{0.83}{
 \begin{tabular}{r|ccc|ccc|l}
  \hline
 \rowcolor{black!10} Method &  Modality & Backbone  & Epochs & AP$_{ped}$ & AP$_{divider}$ & AP$_{boundary}$ & mAP\\
  \midrule
HDMapNet~\cite{li2022hdmapnet}& C & Efficient-B0 & 30 & 13.1 & 5.70 & 37.6 & 18.8\\
VectorMapNet~\cite{liu2023vectormapnet}& C & ResNet-50 & 110 & 38.3 & 36.1 &39.2 &37.9\\
MapTR$^{\dag}$~\cite{MapTR}& C & ResNet-50 & 6 & 58.7 & 59.3 & 60.3 & 59.4\\
MapTR$^{\dag}$~\cite{MapTR}& C \& L & ResNet-50 \& SECOND & 6 & 65.1 & 61.6 & 75.1 & 67.3 \\ 
\hline 
\rowcolor{blue!10} \textbf{MapFusion (Ours)} & C \& L & ResNet-50 \& SECOND & 6 & \textbf{69.4} &\textbf{65.8} & \textbf{78.9} & \textbf{71.4}$_{+4.1}$\\
  \bottomrule
  \end{tabular}}
\end{center}
\end{table*}

\subsection{Comparison with the State-of-the-Arts}
\subsubsection{HD map construction task}
We compare MapFusion with state-of-the-art HD map construction methods on nuScenes  and Argoverse2 datasets.
Our proposed MapFusion outperforms all existing approaches in both single-class APs and the overall mAP by a
significant margin.

\textbf{Experimental Settings.}
We adopt average precision (AP) to evaluate the map construction quality.
Chamfer distance $D_{Chamfer}$ is
used to determine whether the prediction and GT are matched or not. 
We calculate the $AP_\tau$ under
several $D_{Chamfer}$ thresholds ($\tau \in T, T=\{0.5,1.0,1.5\} $, unit is meter), and then average across all thresholds as the final AP metric.
The resolution of source images is $1,600\times900$. 
During the training phase, we resize the source images using a ratio of $0.5$. 
Moreover, we set the maximum number of map elements in one frame, the number of points in one map element, the size of each BEV grid, and the number of transformer decoder layers to $100$, $20$, $0.75\textrm{m}$, and $2$, respectively. 
We follow the experimental settings of existing methods from MapTR~\cite{MapTR}.

\textbf{Experimental Results.}
With the same settings and data partition, we compare the proposed MapFusion method with several state-of-the-art methods, i.e., HDMapNet~\cite{li2022hdmapnet}, VectorMapNet~\cite{liu2023vectormapnet} and MapTR~\cite{MapTR}.
Tab.~\ref{tab1} and Tab.~\ref{tab2}  show the overall performance of MapFusion and all the baselines on nuScenes  and Argoverse2 datasets, respectively. 
Note that re-implementation is needed because the reference methods do not report results on Argoverse2 data set, which has different input data format from nuScenes.
The experimental results reveal a number of interesting points:
(1) The performance of multi-modal methods are obviously better than that of single-modal methods, which proves the significance of utilizing complementary cues from camera and LiDAR to improve the HD map construction performance. 
(2) In the multi-modality setting, the proposed MapFusion approach achieves a 3.6\% absolute improvement in mAP over the previous state-of-the-art MapTR~\cite{MapTR} on the nuScenes dataset. Similarly, it shows a 4.1\% absolute improvement in mAP compared to MapTR~\cite{MapTR} on the Argoverse2 dataset. This advantage arises from the limitations of the three compared HD map construction methods—HDMapNet~\cite{li2022hdmapnet}, VectorMapNet~\cite{liu2023vectormapnet}, and MapTR~\cite{MapTR}—which rely on straightforward channel concatenation and convolution for multi-modal feature fusion, as shown in Fig.~\ref{fig4} (a) (Conv Fusion). These methods neglect modality interaction and employ overly simplistic fusion strategies, resulting in misalignment and information loss.

In a nutshell, MapFusion shows significant superiority over other multi-modal methods, indicating the benefit of cross-modal interaction transform (CIT) module and dual dynamic fusion (DDF) module.
This is due to the fact that the CIT module enables the two feature spaces to interact with each other and enhances feature representation through a self-attention mechanism, while the DDF module automatically selects valuable information from different modalities and can make full use of the inherent complementary information between different modalities.

\begin{table*}
\begin{center}
\caption{\textbf{Results of the BEV map segmentation task on the nuScenes dataset.}
We compare with existing methods from literature, where the numbers are taken from BEVFusion~\cite{liu2023bevfusion}.
We also provide information on the backbones and input modalities in the table. 
MapFusion outperforms the state-of-the-art multi-sensor fusion methods and achieves consistent improvements across different categories.
\textbf{Note that, we use BEVFusion~\cite{liu2023bevfusion} as the baseline model.}
}
\label{tab3}
\scalebox{0.83}{
  \begin{tabular}{r|cc|cccccc|l}
  \hline
\rowcolor{black!10}  Method &  Modality & Backbone & Drivable& Ped. Cross. & Walkway &Stop Line & Carpark &Divider & mIoU\\
 \midrule
 OFT~\cite{roddick2018orthographic} & C & ResNet18 & 74.0 & 35.3 & 45.9 & 27.5 &35.9 &33.9 & 42.1\\
 LSS~\cite{202eecvlls} & C & ResNet18  & 75.4 & 38.8 & 46.3 & 30.3 & 39.1 & 36.5 & 44.4\\
 CVT~\cite{zhou2022cross} & C & EfficientNet-B4 & 74.3 & 36.8 & 39.9 &  25.8 & 35.0 & 29.4 & 40.2\\
 M$^{2}$BEV~\cite{xie2022m} & C & ResNet101 & 77.2 & \XSolidBrush &  \XSolidBrush & \XSolidBrush & \XSolidBrush & $40.5$ & \XSolidBrush\\
 BEVFusion~\cite{liu2023bevfusion} & C & Swin-T & 81.7 &  54.8  & 58.4 & 47.4 & 50.7 & 46.4 & 56.6\\
 X-Align~\cite{borse2023x}& C & Swin-T & 82.4 & 55.6 & 59.3 &49.6 & 53.8 &47.4 &58.0\\
  
\hline

PointPillars~\cite{lang2019pointpillars}& L & VoxelNet & 72.0 &  43.1  &  53.1 & 29.7 &  27.7 & 37.5& 43.8\\
CenterPoint~\cite{yin2021center}&L & VoxelNet & 75.6 &  48.4 &  57.5 &  36.5 & 31.7 & 41.9 & 48.6\\

\hline
PointPainting~\cite{2020cvprpointpainting} & C $\&$ L & ResNet-101 \& PointPillars & 75.9 & 48.5  & 57.1 & 36.9 & 34.5 & 41.9 & 49.1 \\ 
MVP~\cite{21nipsmultimodal} & C \& L & ResNet-101 $\&$ VoxelNet & 76.1 & 48.7 &  57.0 &   36.9 & 33.0 &  42.2 &49.0\\ 
BEVFusion~\cite{liu2023bevfusion}& C \& L & Swin-T $\&$ VoxelNet & 85.5 &  60.5 &  67.6 &  52.0 & 57.0 &  53.7 &62.7\\
 
\rowcolor{gray!20} X-Align~\cite{borse2023x}& C \& L & Swin-T \& VoxelNet &86.8 &  65.2 &70.0 &  58.3 &  57.1 & 58.2 & 65.7\\ 

\rowcolor{blue!10}  \textbf{MapFusion (Ours)} & C $\&$ L & Swin-T $\&$  VoxelNet & \textbf{88.9} & \textbf{69.6} & \textbf{74.0} & \textbf{63.0} & \textbf{56.5} & \textbf{61.5} & \textbf{68.9}$_{+6.2}$\\
  \bottomrule
  \end{tabular}}
\end{center}
\vspace{-1em}
\end{table*}

\begin{table*}
\begin{center}
\caption{\textbf{Results of the BEV map segmentation task on the Argoverse2 dataset.}
 $\dag$  
 denotes our re-implementation following the setting in the paper.} \label{tab4}
\scalebox{0.95}{
 \begin{tabular}{r|cc|ccc|l}
  \hline
\rowcolor{black!10}  Method &  Modality & Backbone & Drivable & $Ped. Cross.$ & Divider & mIoU\\
  \midrule
BEVFusion$\dag$  ~\cite{liu2023bevfusion} & C $\&$ L& Swin-T $\&$  VoxelNet & 78.1 & 30.7 & 46.3 & 51.7\\
\hline
\rowcolor{blue!10} \textbf{MapFusion (Ours)} & C $\&$ L & Swin-T $\&$ VoxelNet & \textbf{83.5} & \textbf{37.4} & \textbf{53.7} & \textbf{58.2}$_{+6.5}$\\
  \bottomrule
  \end{tabular}}
\end{center}
\vspace{-1em}
\end{table*}

\subsubsection{BEV map segmentation task}
We further compare MapFusion with state-of-the-art BEV map segmentation models, where MapFusion outperforms the state-of-the-art multi-sensor fusion methods and achieves consistent improvements across different categories.

\textbf{Experimental Settings.}
We report the Intersection-over-Union (IoU) on $6$ background classes (drivable space, pedestrian crossing, walkway, stop line, car-parking area, and lane divider) on nuScenes dataset and $3$ background classes (drivable space, pedestrian crossing and lane divider) on Argoverse2 dataset. 
The class-averaged mean IoU as our evaluation metric.
For each frame, we only perform the evaluation in the $[-50\mathrm{m}, 50\mathrm{m}]\times[-50\mathrm{m}, 50\mathrm{m}]$ region around the ego car following~\cite{2020cvprpointpainting,21nipsmultimodal,liu2023bevfusion,bevfusion22mips}.
In
MapFusion model, we use a single model that jointly performs binary segmentation for all classes instead
of following the conventional approach to train a separate model for each class. 
We follow the experimental results of existing methods from BEVFusion~\cite{liu2023bevfusion}.

\textbf{Experimental Results.}
With the same settings and data partition, we compare the proposed MapFusion method with several state-of-the-art methods, i.e., PointPainting~\cite{2020cvprpointpainting}, MVP~\cite{21nipsmultimodal}, and BEVFusion~\cite{liu2023bevfusion}.
Tab.~\ref{tab3} and Tab.~\ref{tab4}  show the overall performance of MapFusion and all the baselines on nuScenes and Argoverse2 datasets, respectively. 
Similar to HD map construction, we also re-implemented the experimental results for the Argoverse2 dataset.

The experimental results reveal several interesting points:
(1) In the single-modality setting, camera-based models perform significantly better than LiDAR-based models.
This observation is the exact opposite of results in 3D object detection task~\cite{bevfusion22mips,liu2023bevfusion}.
The main reason is that the map construction task is a semantic-oriented task, which pays more attention to the semantic information in the image.
Therefore, the performance of directly using the fusion method on the 3D object detection task for the map task is not
satisfactory.
(2) In the multi-modality setting, MapFusion outperforms existing state-of-the-art multi-sensor fusion methods, consistently across various categories. This advantage arises from the limitations of these methods: PointPainting~\cite{2020cvprpointpainting} is object-centric, focusing solely on enhancing foreground LiDAR points, while MVP~\cite{21nipsmultimodal} is geometry-oriented, concentrating exclusively on densifying foreground 3D objects—neither effectively segments map components.
Furthermore, BEVFusion~\cite{liu2023bevfusion} and X-Align~\cite{borse2023x} neglect modality interactions and rely on overly simplistic fusion strategies (see Fig.~\ref{fig4}(a) Conv Fusion and Fig.~\ref{fig4}(c) Dynamic Fusion), resulting in misalignment and information loss. Our proposed MapFusion approach achieves a 6.2\% absolute improvement in mean Intersection over Union (mIoU) compared to the previous state-of-the-art BEVFusion \cite{liu2023bevfusion} on the nuScenes dataset, and a 6.5\% absolute improvement on the Argoverse2 dataset.
Notably, we re-implemented the BEVFusion method following the original settings outlined in their paper. Overall, MapFusion consistently enhances the performance of existing fusion methods on both the nuScenes and Argoverse2 datasets, demonstrating the effectiveness of our proposed CIT and DDF components.

\begin{table*}
\begin{center}
\caption{\textbf{An ablation study of the proposed MapFusion components is performed on the nuScenes dataset HD map construction task.} 
``DDF'' and ``CIT'' respectively denote Dual Dynamic Fusion module and Cross-modal Interaction Transform module.
We show the effects of our proposed modules.}
\scalebox{0.9}{
\begin{tabular}{cc|ccc|ccc|l}  
  \hline
 \rowcolor{black!10} DDF & CIT & Modality & Backbone & Epochs & AP$_{ped}$ & AP$_{divider}$ & AP$_{boundary}$ & mAP\\
  \midrule
\XSolidBrush & \XSolidBrush & C \& L & ResNet-50 \& SECOND  & 24 & 55.9 & 62.3 & 69.3 &62.5\\
\CheckmarkBold & \XSolidBrush & C \& L & ResNet-50 \& SECOND & 24 & 58.4 & 64.1 & 72.5 & 65.0$_{+2.5}$\\
\XSolidBrush & \CheckmarkBold & C \& L & ResNet-50 \& SECOND & 24 & 60.2 &64.3 & 72.1 & 65.5$_{+3.0}$\\
\rowcolor{blue!10} \CheckmarkBold & \CheckmarkBold & C \& L & ResNet-50 \& SECOND & 24 & \textbf{61.6}  & \textbf{64.4} & \textbf{72.5} & \textbf{66.1}$_{+3.6}$\\
  \bottomrule  
\end{tabular}}
\label{tab5}
\end{center}
\end{table*}

\begin{table*}
\begin{center}
\caption{\textbf{An ablation study of the proposed MapFusion components is performed on the nuScenes dataset BEV map segmentation task.}
}
\label{tab6}
\scalebox{0.9}{
\begin{tabular}{cc|cc|cccccc|l}
  \hline
\rowcolor{black!10}  DDF & CIT & Modality & Backbone & Drivable & Ped. Cross. & Walkway & Stop Line & Carpark & Divider & mIoU\\
  \midrule
  
   \XSolidBrush &\XSolidBrush & C \& L & ResNet-50 \&  VoxelNet  & 85.5 & 60.5 &  67.6 & 52.0 & 57.0 & 53.7 &62.7\\ 
   
\CheckmarkBold & \XSolidBrush & C \& L & ResNet-50 \&  VoxelNet  & 86.2 & 62.2 & 68.9&   54.4 &56.4&56.0&64.1$_{+1.4}$\\
   
    \XSolidBrush & \CheckmarkBold & C \& L & ResNet-50 \&  VoxelNet  &88.8 & 68.3 & 73.6 & 62.6 & 56.0 & 60.5 & 68.3$_{+5.6}$\\
   
 \rowcolor{blue!10}  \CheckmarkBold & \CheckmarkBold & C \& L & ResNet-50 \&  VoxelNet 
   & \textbf{88.9} & \textbf{69.6}  & \textbf{74.0} & \textbf{63.0} & \textbf{56.5} & \textbf{61.5} & \textbf{68.9}$_{+6.2}$\\
  \bottomrule
  \end{tabular}}
\end{center}
\end{table*}

\subsection{Ablation Studies}

\subsubsection{Contribution of each component}
To systematically evaluate the effectiveness of each module of our proposed MapFusion, we train the model using different components and show the experimental results of the HD map construction task and BEV map segmentation task in Tab.~\ref{tab5} and Tab.~\ref{tab6} respectively.
In the main ablation study, we design the following model variants:
(1) MapFusion (Baseline) : we train the model without the cross-modal interaction transform module and dual dynamic fusion module.
(2) MapFusion (w/ DDF) : we train the model with the dual dynamic fusion module.
(3) MapFusion (w/ CIT) : we train the model with the cross-modal interaction transform module.
(4) MapFusion (full) : we train the model with the cross-modal interaction transform module and dual dynamic fusion module.

The experimental results reveal some interesting findings:
(1) The results of both MapFusion (w/ DDF)  and MapFusion (w/ CIT)  are significantly better than the MapFusion (Baseline), verifying the effectiveness of CIT and DDF components for improving  multi-modal BEV map construction.
Compared with the baseline model,  DDF and CIT modules achieve $2.5$\% and $3.0$\% absolute improvements respectively on HD map construction task, demonstrating the superiority of our approach.
Similarly, compared with the baseline model,  DDF and CIT modules achieve $1.4$\% and $5.6$\% absolute improvements respectively on BEV map segmentation task. 
(2) The results of MapFusion (w/ DDF) and MapFusion (w/ CIT) are inferior to the MapFusion (full), verifying the effectiveness of using both CIT and DDF simultaneously.
MapFusion (full) achieves $3.6$\% and $6.2$\% absolute improvements on the HD map construction and semantic map construction tasks, respectively, demonstrating the superiority of our method.

These experimental results demonstrate that the CIT module enables the camera and LiDAR BEV space to interact with each other to enhance feature representation through the cross-attention mechanism. 
Moreover, it is verified that the DDF module can automatically select valuable information from different modalities, thereby making full use of the inherent complementary information between different modalities.


\begin{table*}
\begin{center}
\caption{
\textbf{Performance comparison of different fusion strategies on HD map construction task.}
Our proposed dual dynamic fusion strategy outperforms all existing approaches by a significant margin.
}
\scalebox{0.83}{
\begin{tabular}{r|ccc|ccc|l}  
  
  \hline
 \rowcolor{black!10}  Method & Modality &  Backbone &  Epochs & AP$_{ped}$ & AP$_{divider}$ & AP$_{boundary}$ & mAP\\
  \midrule
  Baseline(Conv. Fusion) &  C \& L & ResNet-50 \& SECOND & 24 & 55.9  & 62.3 &  69.3 & 62.5\\
  \hline
  \hline
Add Fusion &  C \& L & ResNet-50 \& SECOND & 24 &  \textbf{61.1}  & 60.3 & 71.8 & 64.4$_{+1.9}$\\

Dynamic Fusion & C \& L & ResNet-50 \& SECOND & 24 & 58.4 & 63.1 & 71.5 & 64.3$_{+1.8}$\\

\rowcolor{blue!10} Dual Dynamic Fusion &  C \& L &  ResNet-50 \& SECOND  &  24 &  58.4 & \textbf{64.1} & \textbf{72.5} & \textbf{65.0}$_{+2.5}$\\
  \bottomrule
  \end{tabular}}
\label{tab7}
\end{center}
\end{table*}

\begin{table*}
\begin{center}
\caption{\textbf{Performance comparison of different fusion strategies on BEV map segmentation task.}
}
\label{tab8}
\scalebox{0.83}{
\begin{tabular}{r|cc|cccccc|l}
  \hline
 \rowcolor{black!10} 
 Method &Modality&Backbone& Drivable & Ped. Cross. & Walkway & Stop Line & Carpark & Divider & mIoU\\
  \midrule
Baseline(Conv. Fusion)&C\&L &ResNet-50 \& VoxelNet& 85.5 & 60.5 &  67.6 &  52.0 &  57.0 & 53.7 & 62.7\\ 
\hline
\hline
Add Fusion &C\&L &ResNet-50 \& VoxelNet& 85.4 &60.6 & 67.8 &52.3 & 57.5 & 53.9 & 62.9$_{+0.2}$\\

Dynamic Fusion  & C\&L &ResNet-50 \& VoxelNet& 86.1 &\textbf{62.5} & 68.7 & 53.9 &54.7 & 55.6 & 63.6$_{+0.9}$\\
\rowcolor{blue!10} Dual Dynamic Fusion &C\&L &ResNet-50 \& VoxelNet& \textbf{86.2} & 62.2 & \textbf{68.9} & \textbf{54.4} & \textbf{56.4} & \textbf{56.0} & \textbf{64.1}$_{+1.4}$\\
  \midrule
  \end{tabular}}
\end{center}
\end{table*}

\begin{table*}
\begin{center}
\caption{
\textbf{Compatibility to other HD map construction methods.}  Adding MapFusion leads to consistent performance boost on nuScenes val set in terms of mAP.  
$\dag$  denotes our re-implementation following the setting in the original papers.
}
\label{tab9}
\scalebox{0.83}{
\begin{tabular}{r|c|cc|cccc|l}  
  
  \hline
 \rowcolor{black!5} Method &  Venue &  Modality & Backbone & Epochs & AP$_{ped}$ & AP$_{divider}$ &AP$_{boundary}$ &mAP\\
  \midrule
  HDMapNet$^{\dag}$~\cite{li2022hdmapnet} & ICRA 22 & C \& L & Efficient-B0 \& PointPillars & 30 & 13.3 & 26.9 & 44.3 & 28.2 \\
   \rowcolor{blue!10} HDMapNet $+$ MapFusion & $-$ & C \& L &Efficient-B0 \& PointPillars &  30 & \textbf{21.1} &\textbf{34.2} & \textbf{52.1} & \textbf{35.8}$_{+7.6}$\\
  \hline
  \hline
   VectorMapNet$^{\dag}$~\cite{liu2023vectormapnet} & ICML 23 & C \& L& ResNet-50 \& PointPillars & 110 & 35.8 & 48.2 & 45.3 & 43.1 \\
   \rowcolor{blue!10} VectorMapNet $+$ MapFusion &$-$ &C \& L & ResNet-50 \& PointPillars & 110 & \textbf{41.1} & \textbf{53.7} & \textbf{50.9} & \textbf{48.6}$_{+5.5}$\\
    \hline
    \hline
     MapTR~\cite{MapTR} & ICLR 23 & C \& L &ResNet-50 \& SECOND & 24 & 55.9 &62.3 & 69.3 & 62.5\\
      \rowcolor{blue!10} MapTR $+$ MapFusion & $-$ & C \& L & ResNet-50 \& SECOND & 24 & \textbf{61.6} & \textbf{64.4} & \textbf{72.5} & \textbf{66.1}$_{+3.6}$\\
  \bottomrule
  \end{tabular}}
\end{center}
\end{table*}

\subsubsection{Analysis on different Fusion methods}

To systematically evaluate the effectiveness of the dual dynamic fusion (DDF) method, we train the model using different fusion methods detailed in Section~\ref{sec3.4}.
Tab.~\ref{tab7} and Tab.~\ref{tab8} show the experimental results on the HD map construction task and BEV map segmentation task using different fusion methods, respectively.
For instance, the proposed DDF method achieves $2.5$\% absolute  improvements compared with Baseline model (Conv. Fusion) method on HD map construction task.
Similarly, DDF method achieves $1.4$\% absolute improvement compared with Baseline model (Conv. Fusion) on BEV map segmentation task.
Experimental results show that the DDF module plays a vital role in multi-modal BEV feature fusion and can automatically select valuable information from different modalities for better feature fusion.

\subsubsection{Compatibility with other HD Map Construction methods}
We show MapFusion is compatibility with other HD Map Construction methods, i.e., HDMapNet~\cite{li2022hdmapnet}, VectorMapNet~\cite{liu2023vectormapnet}, and MapTR~\cite{MapTR}.
Besides adding MapFusion, we do not modify their original training settings. 
For all experiments, we report the result of the nuScenes val set.
As shown in Tab.~\ref{tab9}, simply adding MapFusion on top of these strong baselines consistently improves state-of-the-art performance. 
MapFusion demonstrates a significant accuracy boost (absolute): HDMapNet(+$7.6$\%), VectorMapNet (+$5.5$\%), and MapTR (+$3.6$\%). 
This shows the versatility of MapFusion as a multi-modal BEV feature fusion method.

\subsubsection{Accuracy-Computation Analysis}
\begin{figure}[t]
	\setlength{\abovecaptionskip}{-0.0001em}
	\begin{center}
		\includegraphics[width=1.0\linewidth]{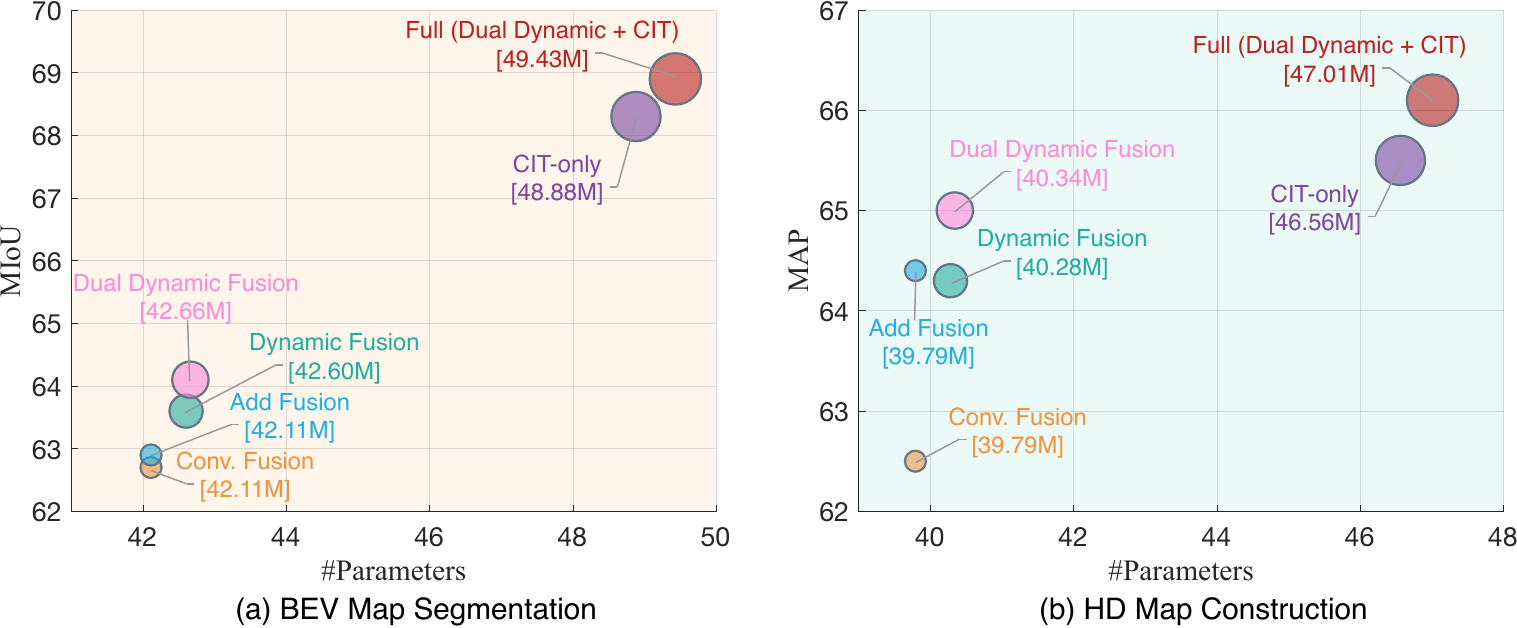}
	\end{center}
	\caption{
\textbf{Accuracy-Computation Analysis.}
We report the accuracy-computation trade-off
by utilizing our proposed cross-modal interaction transform module and different fusion strategies.
	}
	\label{fig5}	
    \vspace{-1em}
\end{figure}

In Fig.~\ref{fig5}, we report the accuracy-computation trade-off
by utilizing our proposed CIT module (See section~\ref{sec3.3}) and different fusion strategies (See section~\ref{sec3.4}).
It can be seen that when using the CIT module, we achieve the highest accuracy improvement at a higher computational cost, while the DDF module introduces less additional cost but provides less performance gain.
It can be seen that all our proposed fusion modules achieve better trade-offs compared with the baselines.
Furthermore, we find that the CIT module significantly outperforms existing BEV fusion strategies, which again verifies that the baseline fusion using simple concatenation and convolutions does not provide the suitable capacity for the model
to align and aggregate multi-modal features.

\begin{figure}[t]
	\setlength{\abovecaptionskip}{-1cm}
	\begin{center}
		\includegraphics[width=1.0\linewidth]{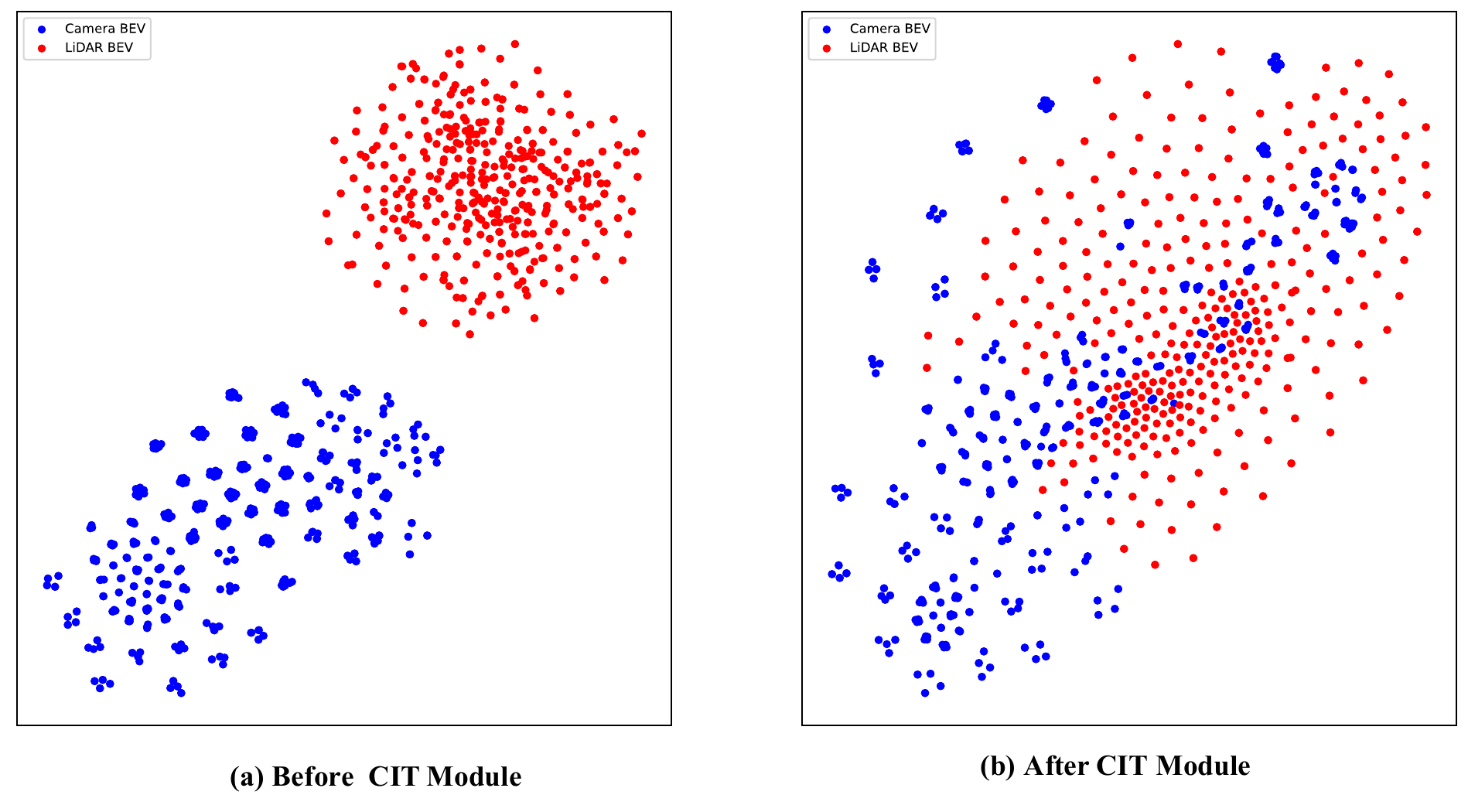}
	\end{center}
	\caption{
The t-SNE visualizations of (a) Before CIT module and (b) After CIT module on HD map construction task.
Red/Blue denotes camera BEV feature/LiDAR BEV feature.
After the CIT module, the BEV features from different modalities are aligned in a shared space, \textit{i.e}, red and blue dots are close after the CIT module (best viewed in color).
	}
	\label{tsne}	
    \vspace{-1em}
\end{figure}

\begin{figure}[t]
	\setlength{\abovecaptionskip}{-0.0001em}
	\begin{center}
		\includegraphics[width=0.85\linewidth]{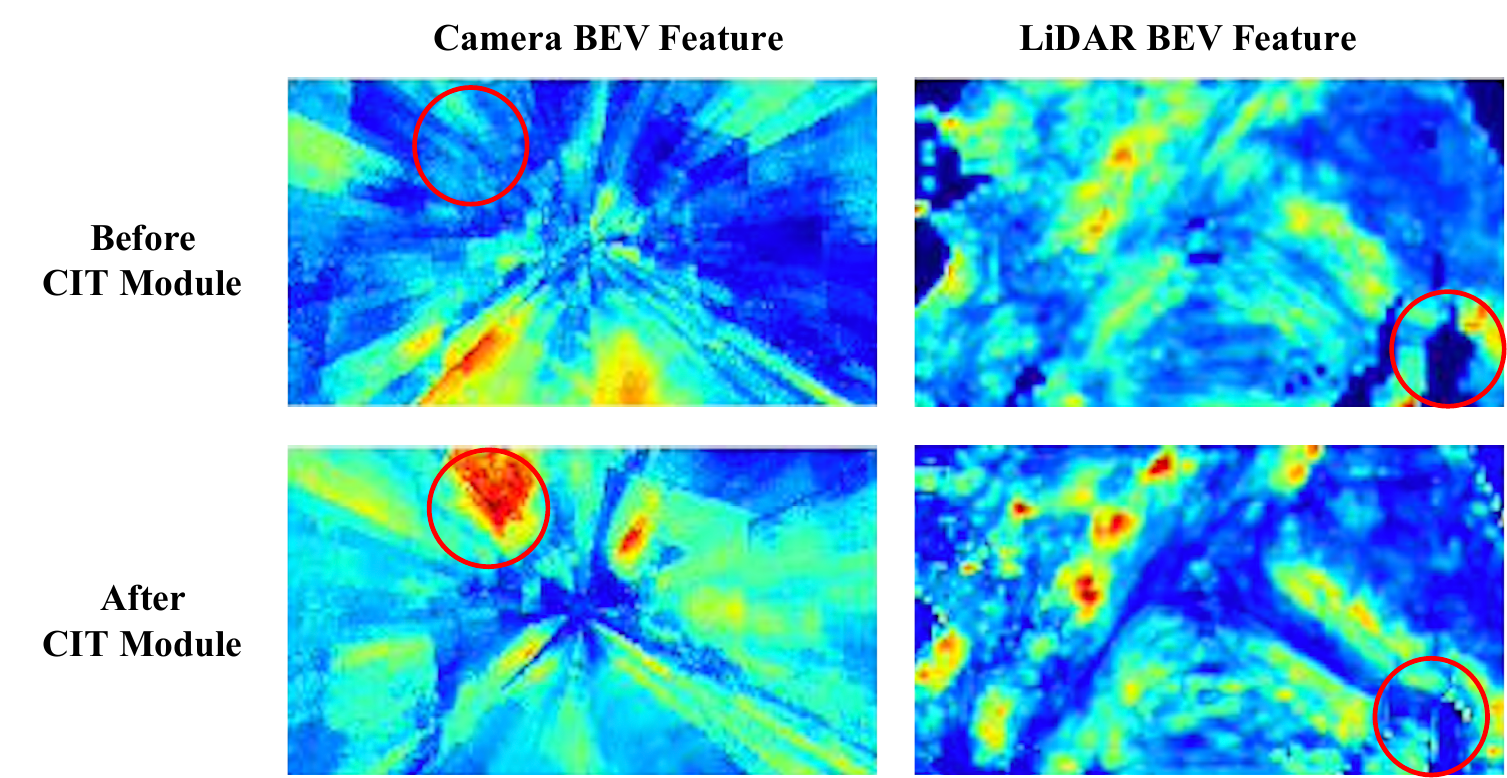}
	\end{center}
	\caption{
Visualization of feature maps before and after the CIT module for the HD map construction task.
	}
	\label{vis}	
    \vspace{-1em}
\end{figure}

\subsection{Visualization}

\textbf{t-SNE.} We randomly choose 500 samples from the nuScenes validation dataset and show the t-SNE~\cite{van2008visualizing} visualizations of (a) Before CIT module and (b) After CIT module in Fig.~\ref{tsne}.
Red/Blue denotes camera BEV feature/LiDAR BEV feature.
As can be seen, Fig.~\ref{tsne} (a) Before CIT module shows that blue and red features are clearly separated, indicating that although in the same space, camera BEV features and LiDAR BEV features can still be misaligned to some extent due to the inaccurate depth in the view transformer and the large modality gap.
Fig.~\ref{tsne} (b) After the CIT module, the BEV features from different modalities are aligned in a shared space, \textit{i.e.}, red and blue dots are close after the CIT module.

\textbf{Feature map visualizations.} In order to visually demonstrate the effectiveness of the CIT module, we visualize the feature map before and after the CIT module in Fig.~\ref{vis}. Before the CIT module, the BEV features of different modalities look quite different. While after the CIT module, they look more similar, verifying mitigated modality misalignment. We can also find that: (1) The camera feature map is enhanced, as shown in the red circles of the left top and left bottom images, making the feature representation more powerful;
(2) Missing features in the LiDAR feature map are recovered, as shown in the red circles of the right top and right bottom images.
In summary, the CIT module integrates different modes of BEV features into a shared space, thereby enhancing representation learning and overall model performance.

\begin{figure*}[!h]
	\centering
 	\includegraphics[width=0.9\textwidth]{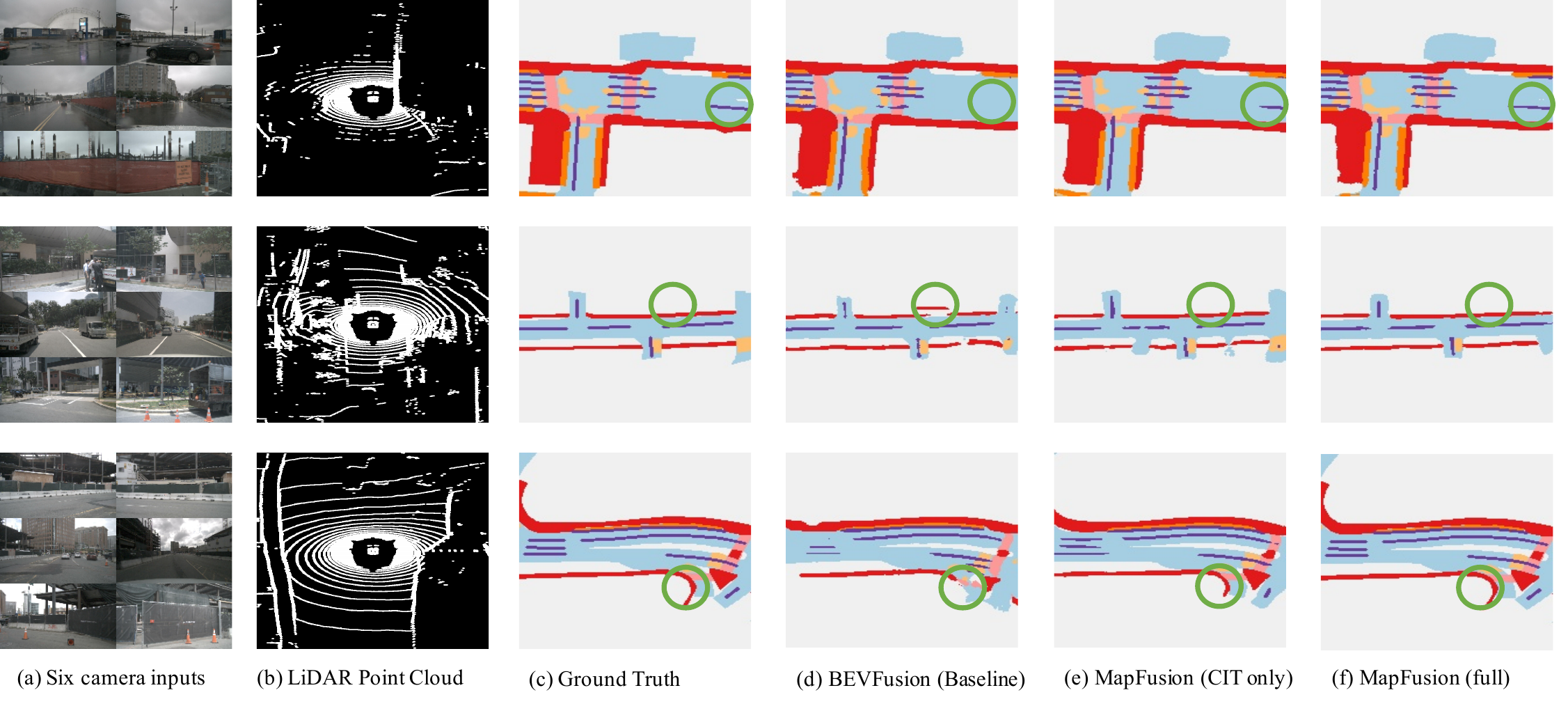}
	\caption{
Qualitative results on BEV map segmentation task. We present a sample scene from nuScenes: a) six camera inputs, b) LiDAR scan, c) ground-truth BEV map segmentation map, d) baseline BEV segmentation map (BEVFusion~\cite{liu2023bevfusion}), e) BEV segmentation map of only using CIT module, and f) BEV segmentation map of MapFusion (full).}
\label{fig6}
\end{figure*}

\begin{figure*}[!h]
	\centering
 	\includegraphics[width=0.9\textwidth]{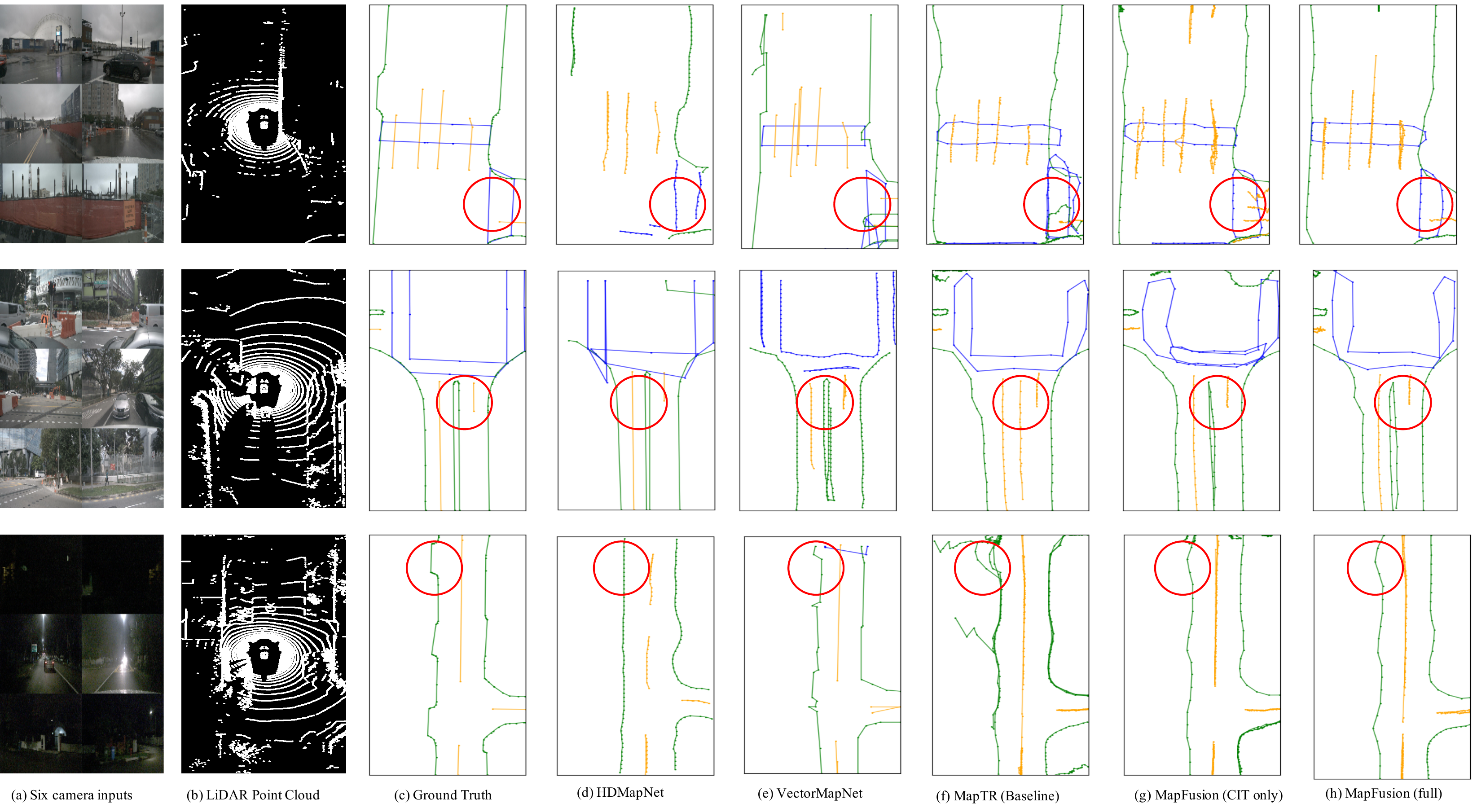}
	\caption{
Qualitative results on HD map task. We present a sample scene from nuScenes: a) six camera inputs, b) LiDAR scan, c) ground-truth BEV vectorized HD map, d) HDMapNet~\cite{li2022hdmapnet}, e) VectorMapNet~\cite{liu2023vectormapnet}, f) baseline BEV vectorized HD map (MapTR~\cite{MapTR}), g) BEV vectorized HD map of only using CIT module, and h) BEV vectorized HD map of MapFusion (full).}
\label{fig7}
\vspace{-0.5em}
\end{figure*}

\textbf{Qualitative Results.}
In Fig.~\ref{fig6}, we present more sample scenes from nuScenes on the BEV map segmentation task. 
Each scene consists of 5 parts: 
a) six surround camera inputs b) LiDAR scan, c) ground-truth BEV segmentation
map, d) baseline BEV segmentation (BEVFusion~\cite{liu2023bevfusion}), e) BEV segmentation using CIT module, and d) BEV segmentation of MapFusion (full).
In Fig.~\ref{fig7}, we present qualitative results on a sample 
scene from nuScenes on the HD map construction task, showing both LiDAR and camera inputs.
We compare the predicted vectorized HD map results of different
models, including HDMapNet~\cite{li2022hdmapnet}, VectorMapNet~\cite{liu2023vectormapnet}, the baseline (MapTR~\cite{MapTR}), MapFusion (only using the
CIT module), and the full MapFusion.
We observe that the baseline model prediction is highly erroneous. By using the CIT module can already correct substantial errors in the baseline prediction, and the full MapFusion model further improves accuracy.
Qualitative results demonstrate the advantages of the CIT module and the DDF module on the multi-modal map construction task.

\section{Conclusion}
To tackle the multi-modal BEV feature fusion problem in multi-modal map construction task, we propose a novel method named MapFusion, which can take advantage of the complementary information between BEV features of different modalities.
Specifically, we first propose Cross-modal Interaction Transform (CIT) module to enhance one modality from another modality by the cross-attention mechanism.
Moreover, we propose a Dual Dynamic Fusion (DDF) module to adaptively select valuable information from two modalities for better feature fusion.
Extensive experiments on several benchmarks demonstrate the superiority of our method.
We also verified the effectiveness of the MapFusion components via an extensive ablation study. 

This paper provides a novel multi-modal BEV feature fusion method MapFusion for optimal fusion of RGB and LiDAR information. 
As shown in our experiments, our MapFusion brings consistent accuracy improvements for two different types of map reconstruction tasks in different datasets. 
Our MapFusion model can be simply integrated into existing pipelines in plug-and-play manner. 
Besides the map reconstruction task, we believe that MapFusion can also benefit other multi-modal perception tasks, which we leave for future work. 

\section*{Acknowledgments}
This work was supported by the National Natural Science Foundation of China (No. 62302139), the National Natural Science Foundation of China (No. 62106259), the National Natural Science Foundation of China (No. 62176081) and the FRFCU-HFUT(JZ2023HGTA0202, JZ2023HGQA0101).

\clearpage

\clearpage

\bibliographystyle{cas-model2-names}

\bibliography{cas-refs}

\end{document}